\documentclass[authoryear,english]{elsarticle}
\usepackage{setspace}
\usepackage{todonotes}
\usepackage{changepage}
\usepackage{amsmath}
\usepackage{pdflscape}
\usepackage{amssymb}
\usepackage{tabularx}
\usepackage{titlesec}
\usepackage{booktabs}
\usepackage{caption}
\usepackage{xcolor}
\usepackage{mathtools}
\usepackage[hyphens]{url}
\usepackage{ragged2e}
\usepackage{pgfplotstable}
\usepackage{longtable}
\usepackage{subcaption}
\usepackage{chngcntr}
\usepackage[margin=3cm]{geometry}
\usepackage{hyperref}
\usepackage{makecell}
\usepackage{multirow}
\usepackage{pifont}
\usepackage{arydshln}
\usepackage{siunitx}
\usepackage{placeins}
\usepackage[figuresright]{rotating}
\usepackage{setspace}
\newcommand{\xmark}{\ding{55}}%

\hypersetup{
    colorlinks=true,
    linkcolor=blue,
    filecolor=blue,      
    urlcolor=blue,
    citecolor=blue,
    breaklinks=true
}

\makeatletter
\def\@author#1{\g@addto@macro\elsauthors{\normalsize%
		\def\baselinestretch{1}%
		\upshape\authorsep#1\unskip\textsuperscript{%
			\ifx\@fnmark\@empty\else\unskip\sep\@fnmark\let\sep=,\fi
			\ifx\@corref\@empty\else\unskip\sep\@corref\let\sep=,\fi
		}%
		\def\authorsep{\unskip,\space}%
		\global\let\@fnmark\@empty
		\global\let\@corref\@empty  
		\global\let\sep\@empty}%
	\@eadauthor={#1}
}
\makeatother

\titleformat{\paragraph}
{\normalfont\normalsize\itshape}{\theparagraph}{1em}{}
\titlespacing*{\paragraph}
{0pt}{3.25ex plus 1ex minus .2ex}{1.5ex plus .2ex}

\titleformat{\subparagraph}
{\normalfont\normalsize\itshape}{\thesubparagraph}{1em}{}
\titlespacing*{\subparagraph}
{0pt}{3.25ex plus 1ex minus .2ex}{1.5ex plus .2ex}

\begin{document}

\begin{frontmatter}	
\title{Global Models for Time Series Forecasting: A Simulation Study}

\author[monash]{Hansika Hewamalage\corref{cor1}}
\author[monash]{Christoph Bergmeir}
\author[unimelb]{Kasun Bandara}

\address{Hansika.Hewamalage@monash.edu, Christoph.Bergmeir@monash.edu, kasun.bandara@unimelb.edu.au}
\address[monash]{Dept of Data Science and AI, Faculty of IT, Monash University, Melbourne, Australia.}
\address[unimelb]{School of Computing and Information Systems, Melbourne Centre for Data Science, University of Melbourne, Melbourne, Australia.}

\cortext[cor1]{Corresponding Author Name: Hansika Hewamalage, Affiliation: Dept of Data Science and AI, Faculty of IT, Monash University, Melbourne, Australia, Postal Address: Faculty of Information Technology, P.O. Box 63 Monash University, Victoria 3800, Australia, E-mail address: Hansika.Hewamalage@monash.edu}

\begin{abstract}
In the current context of Big Data, the nature of many forecasting problems has changed from predicting isolated time series to predicting many time series from similar sources. This has opened up the opportunity to develop competitive global forecasting models that simultaneously learn from many time series, which have shown recently promising results in forecasting competitions.
%
Nevertheless, it still remains unclear under which circumstances global forecasting models can outperform the univariate benchmarks, especially along the dimensions of the homogeneity/heterogeneity of series, the complexity of patterns in the series, the complexity of forecasting models, and the lengths/number of series. Our study attempts to address this problem through investigating the effect from these factors, by simulating a number of datasets that have controllable time series characteristics. We simulate time series from simple data generating processes (DGP), such as Auto Regressive (AR) and Seasonal AR, to complex DGPs, such as Chaotic Logistic Map, Self-Exciting Threshold Auto-Regressive, and Mackey-Glass Equations. The data heterogeneity is introduced by mixing time series generated from several DGPs into a single dataset. The lengths and the number of series in the dataset are varied in different scenarios. 
We perform experiments on these datasets using global forecasting models including Recurrent Neural Networks (RNN), Feed-Forward Neural Networks, Pooled Regression (PR) models and Light Gradient Boosting Models (LGBM), and compare their performance against standard statistical univariate forecasting techniques. 
Our experiments demonstrate that when trained as global forecasting models, techniques such as RNNs and LGBMs, which have complex non-linear modelling capabilities, are competitive methods in general under challenging forecasting scenarios such as series having short lengths, datasets with heterogeneous series and having minimal 
prior knowledge of the patterns of the series. 
This makes these techniques promising candidates for forecasting under uncertain situations as opposed to techniques such as PR and AR models, which assume linearity of the underlying data.

\end{abstract}


\end{frontmatter}

\section{Introduction}
\label{sec:introduction}

In many industries, such as retail, energy, ride-share, and tourism, generating accurate forecasts plays a crucial role in business decision-making scenarios. For example, for retailers such as Amazon and Walmart, sales demand forecasting is important as it provides better grounds for optimising their product inventories~\citep{wen2017multihorizon,bandara2019ICONIP}. In the energy sector, demand forecasts are used to determine the fuel allocation, economic dispatch, and others~\citep{Mandal2006}, whereas accurate service demand forecasts across different geographies are essential for industries such as tourism and healthcare~\citep{claveria2014,Bandara2020-en}. In ride-share services such as Uber, accurate prediction of passenger demand during extreme events can help with better resource allocation and with budget planning in advance~\citep{Zhu2017IEEE}.


%
%
%
%
%
%
%

The paradigm in time series forecasting throughout decades has been to treat every time series as an independent dataset. As a result, traditional forecasting techniques are univariate, consider each time series separately and forecast it in isolation. The Exponential Smoothing State Space Model~\citep[ETS, ][]{hyndman2008ets} and Auto-Regressive Integrated Moving Average Model~\citep[ARIMA, ][]{Box1994-km} are the most prominent examples of such methods. 
However, nowadays many companies are collecting large quantities of time series from similar sources routinely, such as sales in retail of thousands of different products, measurements for predictive maintenance across many machines, smart meter data across many households, etc. Though traditional univariate techniques can still deal with forecasting under these circumstances, they leave the huge potential of learning patterns across time series untapped.

Due to this reason, there has been a paradigm shift in forecasting recently, where now a set of series, as opposed to just one series, is seen as a dataset. Then, a global forecasting model~\citep{Januschowski2020ijf} is trained across all the series in the dataset. The global model has the same set of parameters (e.g., the weights if the global model is a neural network) for all the series in contrast to a local model which has a different set of parameters for every individual series~\citep{Januschowski2020ijf}. 
Nowadays, such global models for forecasting are being introduced at fast pace, e.g., in the works of \citet{Wen2017-xz, Flunkert2017-wp, bandara2020clustering}. They are quickly making their way into practice, and have won all recent prestigious forecasting competitions, such as the M4~\citep{makridakis2020m4} and M5 competition~\citep{makridakism5}, and all competitions held in recent years on the Kaggle platform with a forecasting task~\citep{BOJER2020}.


The premise under which these works operate, and how they explain the almost unreasonable effectiveness of global models, is that the series have to be ``related'' in some way~\citep{Flunkert2017-wp, pmlr-v97-wang19k, Rangapuram2018deep, wen2017multihorizon, bandara2020clustering}, so that the global model can extract patterns across the series. However, none of the works attempts to define or analyse the characteristics of such ``relatedness.'' Furthermore, global models seem to work well even in situations where series are clearly unrelated, such as the M4 forecasting competition, whose dataset is a broad mix of un-aligned time series across many different domains. The competition was won by ES-RNN, a global recurrent neural network~\citep{smyl2020esrnn}. 

Thus, the problem of understanding when and why global forecasting models work, is arguably the most important open problem currently in time series forecasting.
The first work to offer explanations in this space is the recent work by \citet{monteromanso2020principles}, which demonstrates theoretically that, no matter how heterogeneous the data are, there always exists a global forecasting model for any dataset that can perform equally well or even better than a collection of local models. Thus, global forecasting models are in theory not in any way more restricted than local ones, and series don't have to be ``related'' for global forecasting models to perform well. However, mere existence of such a global model does not mean it is straightforward to find or construct such a model. 
Instead of considering relatedness, that study focuses on model complexity. Due to exploiting more data, global models can afford to be more complex than local ones while still generalising better. \citet{monteromanso2020principles} then argue that the complexity of global models can be controlled mainly by using 1) more lags as inputs, 2) using non-linear, non-parametetric models, and 3) using data partitioning. Those authors then proceed to confirm and illustrate their findings empirically using real-world datasets.

Despite the insights of the work of \citet{monteromanso2020principles}, many questions in this space are still unanswered, around under which circumstances global forecasting models work, and how model complexity is best introduced for datasets with different amounts of data complexity, data availability, and relatedness.
In order to define and fully control relatedness of the time series in a dataset, we base our work on simulated time series with known Data Generating Processes (DGP). Then, we define relatedness in terms of the parameters of the known DGPs. This also gives us full control over the complexity of the patterns in the data and the availability of data, rather than relying on real-world datasets which may contain a mixture of many different characteristics or none at all. 
We perform a comprehensive experimental study under controlled conditions, to carefully analyse and quantify the interplay between complexity of the DGP, complexity of the model, amount of data available, and the ``relatedness'' of the series.




Our study has the following key contributions. 
We simulate different datasets using a number of carefully chosen DGPs which cover a variety of time series characteristics, most of them closely simulating real-world scenarios. The DGPs that we have chosen vary from simple linear AR(3), to Seasonal Auto-Regressive (SAR) DGPs, to more complex and non-linear DGPs such as Chaotic Logistic Map, Mackey-Glass Equation and Self-Exciting Threshold Auto-Regressive (SETAR) models. The simulation settings that we use in our study along with the respective DGP implementations are available publicly as a code repository\footnote{Currently available at:	\url{https://drive.google.com/file/d/1Qllh4rDEZ26_-fdKTOzi3qAMo3m9tZI_/view?usp=sharing}. A github link will be provided in the final version of the manuscript.}. We have designed several experimental scenarios with each one of these DGPs. To explore the effect of data availability on the model performance, we vary the amount of data in the datasets in terms of both the number of series as well as the lengths of the individual series. For different degrees of data relatedness, we investigate both homogeneous and heterogeneous settings by changing the number of DGPs mixed within the same dataset. Thirdly, we perform experiments with a number of selected global forecasting models including Recurrent Neural Networks (RNN), Pooled Regression (PR) models,
Feed-Forward Neural Networks (FFNN) and Light Gradient Boosting Machines~\citep[LGBM, ][]{lightgbm2017} on the simulated datasets under the different experimental scenarios. The complexity of the different global forecasting models is further changed by introducing two model setups, namely the GFM.All setup where all the series are fitted using the same global model and the GFM.GroupFeature setup, which on top of the GFM.All setup involves extra features to indicate the invidual DGPs the series are generated from within a heterogeneous context. Finally, the performance of the models is compared against each other as well as against the univariate techniques of AR and SETAR models. Based on the empirical evidence, an analysis is provided around the different factors which affect the performance of different global forecasting models.

The rest of the paper is structured as follows. 
Section \ref{sec:overview_of_methodology} first details the overall methodology employed in the study including the different experimental scenarios designed. Next, Section \ref{sec:data_generating_processes} provides a brief review of the different studies which involve time series simulation, followed by the details of the DGPs involved in this study. In Section \ref{sec:forecasting_framework}, we provide the details of the different forecasting techniques that are compared against each other in the study along with their data preprocessing, model training and testing information. Section \ref{sec:experimental_setup} presents the details of the dataset characteristics along with the evaluation metrics as well as the tests for statistical significance of the performance differences between the methods. In Section \ref{sec:results_analysis}, we present an analysis of the results of the experiments as well as a comparison of the computational times of the employed forecasting techniques. Section \ref{sec:conclusions} concludes the paper with the final remarks of the overall study, the conclusions derived and the way forward. We also have an Online Appendix\footnote{\url{https://drive.google.com/file/d/1LDCEJzOmbmTN3wZXfKUyK8cgInbWaONM/view?usp=sharing}} where we include further details of the used DGPs, forecasting techniques, experimental setup and results analysis.

\section{Overview of the Methodology}
\label{sec:overview_of_methodology}
\begin{figure*}[htb]
	\centering
	\includegraphics[scale=0.12]{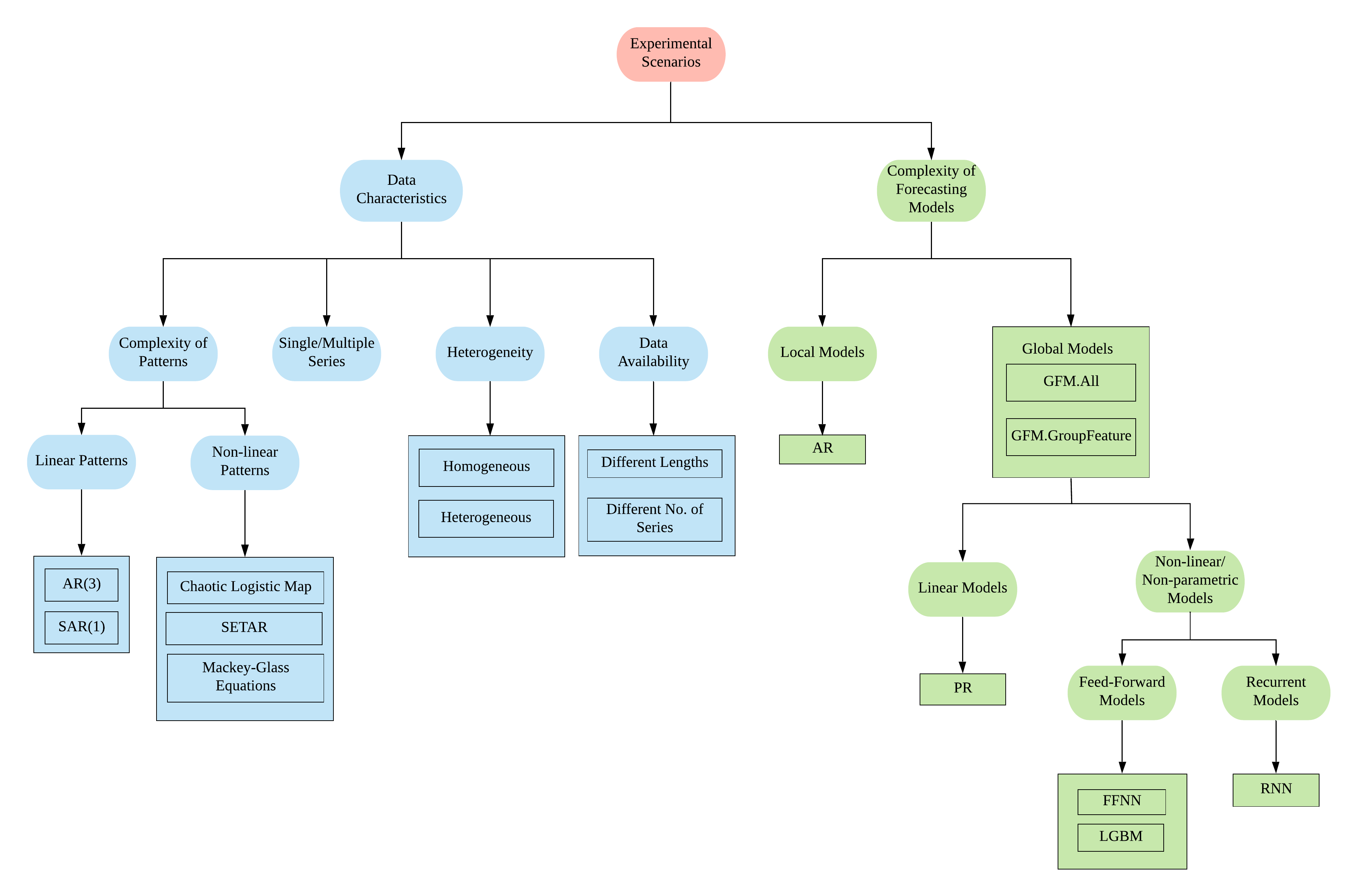}
	\caption{Visualisation of Experimental Settings Employed in the Study}
	\label{fig:simulation_workflow}
\end{figure*}

The methodology of our study involves first simulating a number of datasets having many different characteristics. Once the datasets are simulated as required, different forecasting techniques are tested on them. We introduce a host of experimental scenarios to quantitatively explore the interplay between the complexity of different forecasting models and characteristics of the data in this manner. In our experiments, characteristics of the data are controlled along four dimensions: 1) complexity, 2) single or multiple series, 3) heterogeneity, and 4) amount of data points, whereas forecasting models are controlled using different levels of complexity. The specific details of these controlled experimental settings are explained in this section, and Figure \ref{fig:simulation_workflow} gives an overview.

\subsection{Complexity of the Patterns in the Data}

The complexity of the simulated time series is determined by the complexity of the underlying DGP. The different DGPs that we use for the experiments are as follows. 

\begin{itemize}
	\item AR(3)
	\item SAR(1)
	\item Chaotic Logistic Map
	\item SETAR
	\item Mackey-Glass Equation
\end{itemize}

Out of these DGPs, as discussed in more detail under Section \ref{sec:data_generating_processes}, AR(3) and SAR(1) are linear DGPs whereas the rest generate time series with more complex and non-linear characteristics.

\subsection{Single or Multiple Series}

With every DGP used, series are both used for single series settings and split into multiple equal length series to constitute a scenario with multiple series. The purpose of this experimental scenario is to quantitatively evaluate whether it makes any difference for the models under consideration to learn from the same data either on one long series or across multiple series. For both the scenarios, we evaluate on the same forecast horizon. This means that for the multiple series scenarios, the evaluation is performed on the test forecast horizon of the series that corresponds to the last segment of the relevant long single series. In terms of the forecasting techniques, this indicates that the techniques that can be built as global models can leverage that facility when in the multiple series mode. However, for local forecasting models such as AR and SETAR models, learning is always restricted to use only one series at a time. Consequently, in single time series scenarios, such local models have more training data than in the multiple series scenarios. 

\subsection{Scale of Heterogeneity}

We control the scale of heterogeneity of time series by simulating the following scenarios using above DGPs. 

\begin{enumerate}
	\item Homogeneous (Single) Pattern
	\item Heterogeneous Patterns (Only for Multiple Series)
\end{enumerate}

For multiple series, homogeneity means that all time series are simulated from the same DGP with different seeds. The multiple series scenarios can be made heterogeneous by combining time series generated from different DGPs into a single dataset. Different patterns can also be created using the same DGP by adding Gaussian noise over the coefficients. Thus, the heterogeneity experiments are performed using all DGPs, by generating each simulated time series from a new set of coefficients.

\subsection{Amount of Data Available}
We also control the availability of data by changing the lengths of the individual series or varying the number of series in the dataset. For single series and multiple series scenarios, this is performed using two different setups.

\begin{enumerate}
	\item Single Series Scenarios - Varying the length of the series
	\item Multiple Series Scenarios
	\begin{enumerate}
		\item Homogeneous Patterns - Varying the lengths of the series, Varying the number of series
		\item Heterogeneous Patterns - Varying the lengths of the series
	\end{enumerate}
\end{enumerate} 

For the single series scenario, we can only change the length of the single series. However, for the multiple series scenarios with a single pattern (homogeneous), we simulate both with varying number of series and varying lengths of the series. For the multiple series scenarios with heterogeneous patterns, we only vary the amount of data by changing the lengths of the series. 


Based on the different experimental considerations relevant to each DGP, such as data heterogeneity or single and multiple series inclusion, we introduce the terminology summarised in Table \ref{tab:abbreviations} that is used throughout the rest of this paper.

\begin{table*}
	\footnotesize
	\begin{center}
		\begin{tabular}{lc}
			\toprule
			\textbf{Abbreviation} & \textbf{Meaning}\\
			\hline
			SS & Single Series\\
			\hline
			MS-Hom-Short & Many Series Homogeneous with Short Lengths\\
			\hline
			MS-Hom-Long & Many Series Homogeneous with Long Lengths\\
			\hline
			MS-Het & Many Series Heterogeneous\\
			\hline
		\end{tabular}
		\caption{Explanations of Abbreviations}
		\label{tab:abbreviations}
	\end{center}
\end{table*}

\subsection{Complexity of Forecasting Techniques}
\label{sec:complexity_of_forecasting_techniques}

The modelling capability of models is varied by conducting experiments using a number of techniques with different attributes as explained more under Section \ref{sec:forecasting_framework}.
The complexity of these base models can be further increased by using two techniques. The first is by increasing the number of parameters in the model. For NNs, the hyperparameters are tuned by using automated hyperparameter tuning techniques. However, the size of the input window can be increased as required to improve the complexity. For pooled regression models this can be done by increasing the lag size. The second approach is to introduce different training paradigms for global models as mentioned below. 

\begin{enumerate}
	\item GFM.All: Where the global models are trained using all the series available in the dataset irrespective of their potential heterogeneity. This setup is used in all the multiple series experimental scenarios.
	\item GFM.GroupFeature: Where the global models are trained using all the series available in the dataset similar to the GFM.All setup, but in addition to the time series data, every input has an additional feature to indicate which subgroup the particular series belongs to. This approach is used to attend to existing heterogeneity in the data. It can act as an implicit clustering of the data by giving information to the models of the existence of different subgroups. However, this training paradigm is irrelevant for the heterogeneous setting that we described before, since none of the series in the dataset share coefficients. Therefore, we introduce a separate experimental scenario named Group Feature setup where we mix few patterns from each DGP within a dataset. 
\end{enumerate} 

The GFM.All setup is the base setup. The GFM.GroupFeature setup is expected to increase the complexity of the GFM.All setup.

\section{Data Generating Processes}
\label{sec:data_generating_processes}


Generating synthetic datasets is a heavily used approach in many domains to evaluate the performance of different algorithms, against other benchmarks. Our interest in this study lies specifically around those methods that can generate time series with controllable features, rather than random generation processes. In this section we first mention several related work of time series simulation and then move on to explain the different DGPs used in our study.

\subsection{Background}

Table \ref{tab:dgp_background} summarises the literature related to time series simulation in different domains. For more detailed explanations refer to Appendix A.1 of our Online Appendix. However, most of the work present in the literature builds univariate models for forecasting as opposed to global forecasting models. Most importantly, none of these studies clearly investigates how the simplicity/complexity of the series, complexity of the models, the heterogeneity/homogeneity of the data or the amount of data overall/per series affects the forecasting performance of global models. On the other hand, many of the recent studies involving global forecasting models such as RNNs, claim that global models only work with sets of related time series~\citep{Flunkert2017-wp, Wen2017-xz, bandara2019ICONIP, Januschowski2020ijf}. The study by \citet{bandara2020clustering} uses special clustering mechanisms to group the related time series together to build global models on every cluster. Yet, the notion of relatedness is not well-understood in many of these contexts. No proper efforts have been made to analyse these terms within controlled experimental settings. Therefore, in this study, we strive to explore the effect of these different factors for building global forecasting models in comparison to other state-of-the-art univariate forecasting benchmarks.

\begin{table*}
	\footnotesize
	\begin{center}
		\begin{tabular}{ll}
			\toprule
			\textbf{Domain} & \textbf{Literature}\\
			\hline
			Time Series Classification & \begin{tabular}{@{}l@{}}\citet{SUN201924} \\ \citet{ZHAO2018171}\end{tabular}\\
			\hline
			Time Series Forecasting & \begin{tabular}{@{}l@{}l@{}l@{}}\citet{BERGMEIR2012CV} \\ \citet{bergmeir2018cv} \\ \citet{hyndman2002ets} \\ \citet{Petropoulos2014horses}\end{tabular}\\
			\hline
			Machine Learning & \begin{tabular}{@{}l@{}l@{}l@{}l@{}l@{}l@{}l@{}}\citet{LAU20081539} \\\citet{YE2021107617}\\\citet{VANLI20191}\\\citet{Fischer2018simulation}\\\citet{Zhang2005-le}\\\citet{zhang2007simulation}\\\citet{zhang2001simulation}\\\citet{Suhartono2018simulation}\\ \end{tabular}\\
			\bottomrule
		\end{tabular}
		\caption{Related Work of Time Series Simulation in Different Domains}
		\label{tab:dgp_background}
	\end{center}
\end{table*}

\subsection{Data Generating Processes Used for the Study}

The experimental settings of our study involve scenarios with simple/complex patterns, homogeneous/heterogeneous series, varying number of series, and varying series lengths, as described under Section \ref{sec:overview_of_methodology}. The DGPs that we use for the study are selected such that they can simulate practical forecasting scenarios as closely as possible. We start from the simplest forms of DGPs, which are linear DGPs with short memory. Then, we gradually make them more complex by increasing the number of lags to have linear DGPs with longer memory. Next, we introduce more complex, non-linear DGPs namely, a threshold linear model that switches between different linear models based on certain conditions, a chaotic model and another delay differential equation based model, the Mackey-Glass Equations. This section gives details of these DGPs that we use for our experiments in order to control the complexity of the patterns in the data. Further details corresponding to every DGP can be found under Appendix A.2 of our Online Appendix.

\subsubsection{Linear Auto-Regressive Models}

We first simulate series using AR models. 
ARIMA models attempt to model the autocorrelation in the data~\citep{robgeorg2018otext}. AR models use a linear combination of the past values of the target series itself as regressors to predict the future values. MA models, on the other hand, model the future in terms of a linear combination of the past forecast errors. AR and MA models, once combined together with sufficient differencing of the time series to achieve stationarity, are called ARIMA models. The number of lags used for the AR model (p) and the MA model (q) and the amount of differencing (d) defines the complexity of the underlying ARIMA model. ARMA models can be approximated by pure AR models using many lags~\citep{kang2019gratis}. In this study, we generate series from AR models to enforce linear relationships in the simulated data. For the convenience of explanation, we first introduce the backshift operator as in Equation \ref{eqn:backshift_operator}.

\begin{equation}
\begin{split}
By_{t} &= y_{t-1}\\
B(By_{t}) &= B^{2}y_{t} = y_{t-2}
\label{eqn:backshift_operator}
\end{split}
\end{equation}

The backshift operator on the variable $y_{t}$ is used to denote its lags. 

\paragraph{Non-Seasonal Auto-Regressive Models}

Equation \ref{eqn:ar_model} defines a non-seasonal AR model of order $p$ using the backshift operator.

\begin{equation}
y_{t} = c + \phi_{1}By_{t} + \phi_{2}B^{2}y_{t} + \phi_{3}B^{3}y_{t} + ... + \phi_{p}B^{p}y_{t} + \epsilon_{t}
\label{eqn:ar_model}
\end{equation}

Here, $\epsilon_{t}$ indicates a white noise error term, which is the error in the AR modelling, and $c$ is a constant term which is also known as the intercept. As Equation \ref{eqn:ar_model} shows, an AR model is a simple linear regression model where the past lags are the predictors of the value of the series at the $t^{th}$ time step. Therefore, AR DGPs can simulate those time series common in the real world where the value at a certain time is a (linear) function of its own past values. Given a particular time series, the coefficients $\phi_{1}, \phi_{2}, ..., \phi_{p}$ are estimated to fit an AR model by minimising a loss criterion such as the Maximum Likelihood Estimation (MLE). Similarly, given the values for the coefficients $\phi_{1}, \phi_{2}, ..., \phi_{p}$ of the different lags, the AR model can simulate series. In this study, we use an AR(3) DGP to simulate simple linear patterns in the time series data. We use the \texttt{arima.sim} function from the \texttt{stats} package of the R core libraries~\citep{r_language} to simulate series using AR(3) DGP. 

\paragraph{Seasonal Auto-Regressive Models}

To make the patterns more complex, we also simulate series using an SAR model. Equation \ref{eqn:seasonal_ar_model} shows such an SAR model of order $P$, where the period is denoted by $S$.

\begin{equation}
y_{t} = c + \Phi_{1}B^{S}y_{t} + \Phi_{2}B^{2S}y_{t} + \Phi_{3}B^{3S}y_{t} + ... + \Phi_{P}B^{PS}y_{t} + \epsilon_{t}
\label{eqn:seasonal_ar_model}
\end{equation}

Comparing Equation \ref{eqn:ar_model} with Equation \ref{eqn:seasonal_ar_model}, in the Seasonal AR model, the predictor variables are now seasonal lags as opposed to normal lags as in the non-seasonal AR model. Therefore, the Seasonal AR model is a regression model where the predictor variables are seasonal lags of the time series. Thus, an SAR DGP can simulate time series having a seasonality of a particular periodicity which are also very commonly seen in real-world scenarios. Similar to the non-seasonal AR model, given the values of the coefficients $\Phi_{1}, \Phi_{2}, ..., \Phi_{p}$, the model can simulate new time series. However, unlike in simple AR models, since it is the seasonal lags that are significant in the SAR models, they account for much longer-term dependencies, thus generating series with longer memory. In our study, to generate series with SAR models, we first fit an SAR model of order 1 to the \texttt{USAccDeaths} monthly series of the \texttt{datasets} package~\citep{r_language} available in the R core libraries. Then, we use this model to simulate series using the \texttt{simulate} function of the \texttt{forecast} package~\citep{forecast_package}.

\subsubsection{Chaotic Logistic Map}
\label{sec:chaotic_logistic_map}

The Chaotic Logistic Map DGP is another technique used in this study to generate more complex patterns in the time series data. This type of logistic models were first introduced in biological research to describe the population growth of certain species over time~\citep{May1976}. Chaos theory is used in mathematics to deal with heavily non-linear dynamical systems. Thus, Chaotic Logistic Maps are well-suited techniques for simulating non-linear characteristics in time series which can also be found in the real world. The DGP used in this study generates data as per Equation \ref{eqn:chaotic_logistic_map}. This is a zero-bounded chaotic process where $\epsilon_{t}$ indicates a white noise error term. $r$ is the coefficient of the model.

\begin{equation}
	\label{eqn:chaotic_logistic_map}
	y_t = max\left(r y_{t-1} \left(1-y_{t-1}\right) + \frac{\epsilon_{t}}{10} , 0\right)
\end{equation}


\subsubsection{Self-Exciting Threshold AutoRegressive Models}
\label{sec:setar_dgp}

The SETAR model is also a technique used for simulating complex patterns. This model belongs to the family of TAR models, first introduced by \cite{Tong1978TAR}. A TAR model is defined as in Equation \ref{eqn:tar_model}.

\begin{equation}
y_{t} = 
\begin{cases}
c_{1} + \phi^{1}_{1}By_{t} + \phi^{1}_{2}B^{2}y_{t} + \phi^{1}_{3}B^{3}y_{t} + ... + \phi^{1}_{p}B^{p}y_{t} + \epsilon_{t} ,& \text{if } z_{t} \leq r_{1}\\
c_{2} + \phi^{2}_{1}By_{t} + \phi^{2}_{2}B^{2}y_{t} + \phi^{2}_{3}B^{3}y_{t} + ... + \phi^{2}_{p}B^{p}y_{t} + \epsilon_{t} ,& \text{if } r_{1} \leq z_{t} \leq r_{2}\\
...\\
c_{k} + \phi^{k}_{1}By_{t} + \phi^{k}_{2}B^{2}y_{t} + \phi^{k}_{3}B^{3}y_{t} + ... + \phi^{k}_{p}B^{p}y_{t} + \epsilon_{t} ,& \text{if } r_{k-1} \leq z_{t} \\
\end{cases}
\label{eqn:tar_model}
\end{equation} 

According to Equation \ref{eqn:tar_model}, the TAR model involves $k-1$ number of threshold values $(r_{1}, r_{2}, ..., r_{k-1})$, which separate the space into $k$ regimes, where each one is modelled by a different AR process of order $p$. The threshold variable is denoted by $z_{t}$ which is an exogenous variable. Therefore, in contrast to linear AR models, TAR models capture non-linear dynamics in time series by means of a regime-switching technique that changes the underlying AR coefficients, when a particular threshold value is met. When the threshold variable $z_{t}$ is a lagged value of the series itself (denoted by $y_{t-d}$, where $d$ is the delay parameter), the model is known as a SETAR model. The SETAR models were first introduced by \citet{Tong1980SETAR}. Due to their regime-switching nature, SETAR models can capture, for example, real-world scenarios where patterns of a certain time series change due to policy interventions upon reaching a certain threshold value in the time series. For this study, we use the simulation capability of SETAR models implemented in the \texttt{setar.sim} function of the \texttt{tsDyn} R package~\citep{Fabio2019tsdyn}. 

\subsubsection{Mackey-Glass Equation}

The Mackey-Glass Equation forms another DGP that can be used to simulate complex patterns in time series data. This equation was first introduced in a research studying first-order non-linear Differential-Delay Equations (DDE) which describe physiological control systems~\citep{mackey1977}. The Mackey-Glass Equation was specifically used to explain the fluctuations of white blood cells in the human body under certain cases of chronic leukemia. The solutions of these DDEs can be chaotic and thus cause complexity in the underlying time series. The Mackey-Glass equation is defined as in Equation \ref{eqn:mackey_glass}.

\begin{equation}
	\label{eqn:mackey_glass}
	\frac{dy_t}{dt} = \frac{\beta y_{t-\tau}}{1+y^{n}_{t-\tau}} - \gamma y_t
\end{equation}

In Equation \ref{eqn:mackey_glass}, $\tau$ is called the delay whereas $\beta, \gamma, n > 0$ are parameters that determine the periodicity and the chaos induced into the resulting series. Following from Equation \ref{eqn:mackey_glass}, the solution to the Mackey-Glass equation is as shown in Equation \ref{eqn:mackey_glass_solution}. Therefore, Equation \ref{eqn:mackey_glass_solution} can be directly used to simulate complex time series. For this study we use the Mackey-Glass based time series generation implemented in the \texttt{nolitsa} Python package~\citep{nolitsa}. 

\begin{equation}
	\label{eqn:mackey_glass_solution}
	y_{t+1} = y_t + \frac{\beta y_{t-\tau}}{1+y^n_{t-\tau}} - \gamma y_t
\end{equation}


Table \ref{tab:time_series_simulation_techniques} provides an
overall summary of the aforementioned techniques used for time series simulation. Here, the \textit{Function} and \textit{Package} columns provide the references to the respective software implementations of the simulation techniques used in our experiments. The Chaotic Logistic Map DGP is implemented by ourselves and does not involve any already existing software packages. The table furthermore indicates a characterisation of the generated time series in terms of their linearity.

\begin{table*}
	\footnotesize
	\begin{center}
		\begin{tabular}{lcccc}
			\toprule
			\textbf{Simulation Technique} & \textbf{Function} & \textbf{Package} & \textbf{Linearity}\\
			\hline
			AR(3) & \texttt{arima.sim} & \texttt{stats}~\citep{r_language} &\color{green}\checkmark \\ 
			\hline
			SAR(1)&\makecell{\texttt{simulate}\\ \texttt{sim.ssarima}} & \makecell{\texttt{forecast}~\citep{forecast_package}\\ \texttt{smooth}~\citep{Ivan2019smooth}} & \color{green}\checkmark \\
			\hline
			Chaotic Logistic Map &- &- &\color{red}\xmark\\
			\hline
			SETAR & \texttt{setar.sim} & \texttt{tsDyn}~\citep{Fabio2019tsdyn} & \color{red}\xmark\\
			\hline
			Mackey-Glass Equation & \texttt{data.mackey\_glass} & \texttt{nolitsa}~\citep{nolitsa} & \color{red}\xmark\\
			\hline
		\end{tabular}
		\caption{Summary of Techniques Used for Time Series Simulation}
		\label{tab:time_series_simulation_techniques}
	\end{center}
\end{table*}

\section{Forecasting Framework}
\label{sec:forecasting_framework}


Depending on the specific experimental scenario, we train models either as univariate models or global forecasting models. In this section, the details of the different forecasting techniques employed are explained along with their associated data preprocessing methodologies, model training and testing.

\subsection{Forecasting Techniques}
For the forecasting techniques of the study, we use RNNs, FFNNs, PR models, LGBMs, AR, SETAR and SAR models as appropriate. The details of these techniques are as follows. Further details can be found under Appendix B of our Online Appendix.

\subsubsection{Recurrent Neural Networks}
\label{sec:rnn_architecture}

\begin{figure*}[htbp!]
	\begin{center}
		\includegraphics[scale=0.46]{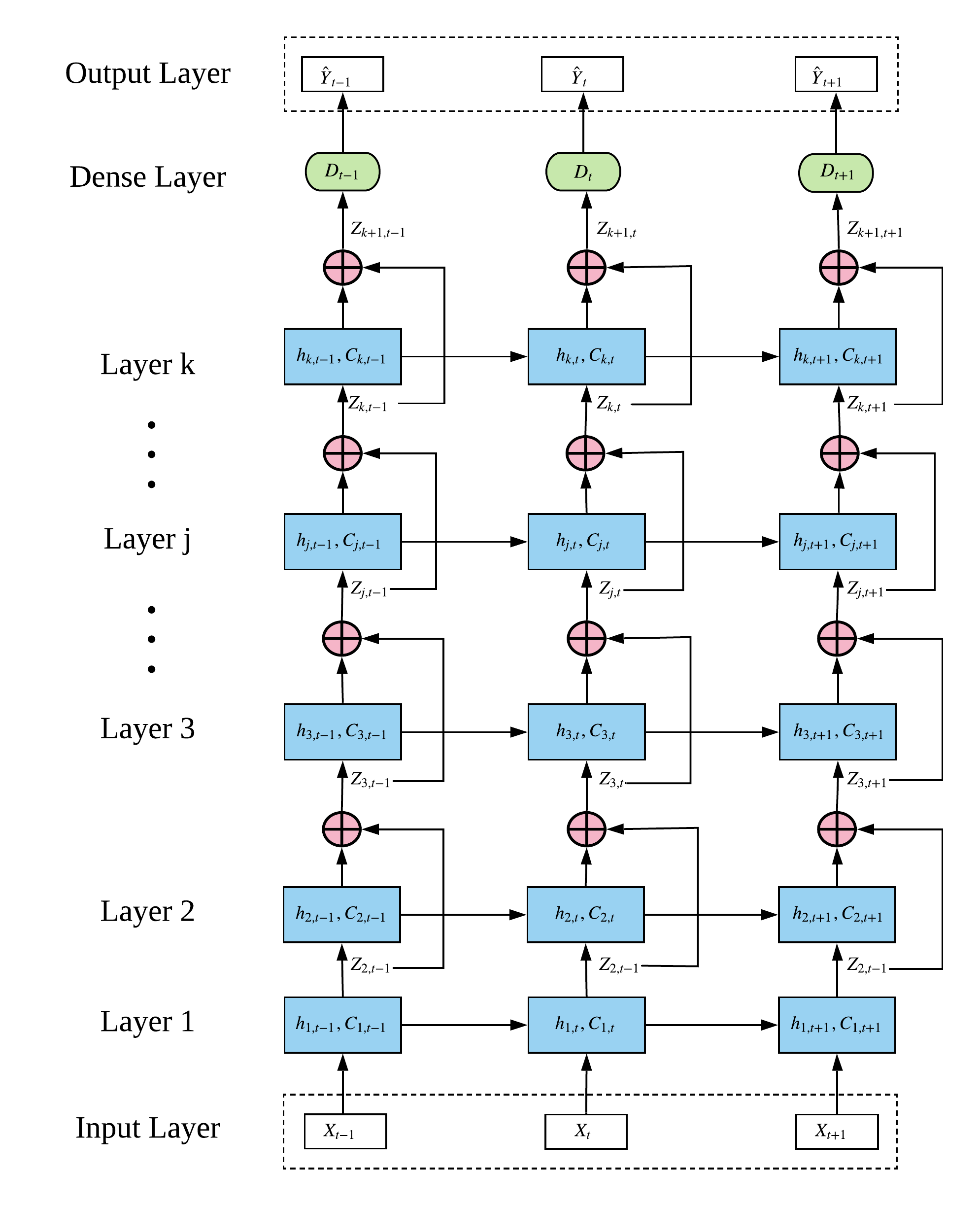}
		\caption{Recurrent Neural Network with Residual Connections}
		\label{fig:residual_rnn}
	\end{center}
\end{figure*}

RNNs are a type of NNs that are specialised for sequence modelling problems due to their states which are propagated up to the end of the sequence and thus help them distinguish between every individual series among a set of series. Like other NNs, RNNs too are univeral approximators meaning that they are inherently non-linear models~\citep{Schafer2006-ln}. RNNs can be implemented with many different architectures. In the study by \cite{hewamalage2019recurrent}, those authors have identified the Stacked architecture as the generally best RNN architecture across a number of real-world datasets. Therefore, in this study we choose that same architecture along with residual connections. Residual connections were introduced by \citet{He2016Resnet} on CNNs mainly for the purpose of image recognition. Such networks were named as Residual Nets (ResNets). However, later many other domains were inspired by this architecture and implemented different modified versions of it for their own work. For the time series forecasting domain, \citet{Smyl2016mcmc} used residual connections on RNN layers. Moreover, the winning solution by \citet{smyl2020esrnn} at the M4 forecasting competition also involved RNNs with residual connections. Inspired by this work, we use an architecture similar to the work of \citet{Smyl2016mcmc} in our study. In particular, our RNN architecture is illustrated in Figure \ref{fig:residual_rnn}. The recurrent unit used for the RNN layers in this study is the Long Short-Term Memory (LSTM) cell with peephole connections introduced by \citet{Gers2003peephole}.
%
The residual connections are especially helpful to avoid vanishing gradient issues of the layers of NNs. 
The RNNs used for this study are implemented with version $2.0$ of the Tensorflow open-source deep learning framework~\citep{tensorflow2015-whitepaper}.

\subsubsection{Feed-Forward Neural Networks}
\label{sec:ffnn_architecture}

\begin{figure*}[htbp!]
	\begin{center}
	\hspace*{-1cm}
		\includegraphics[scale=0.45]{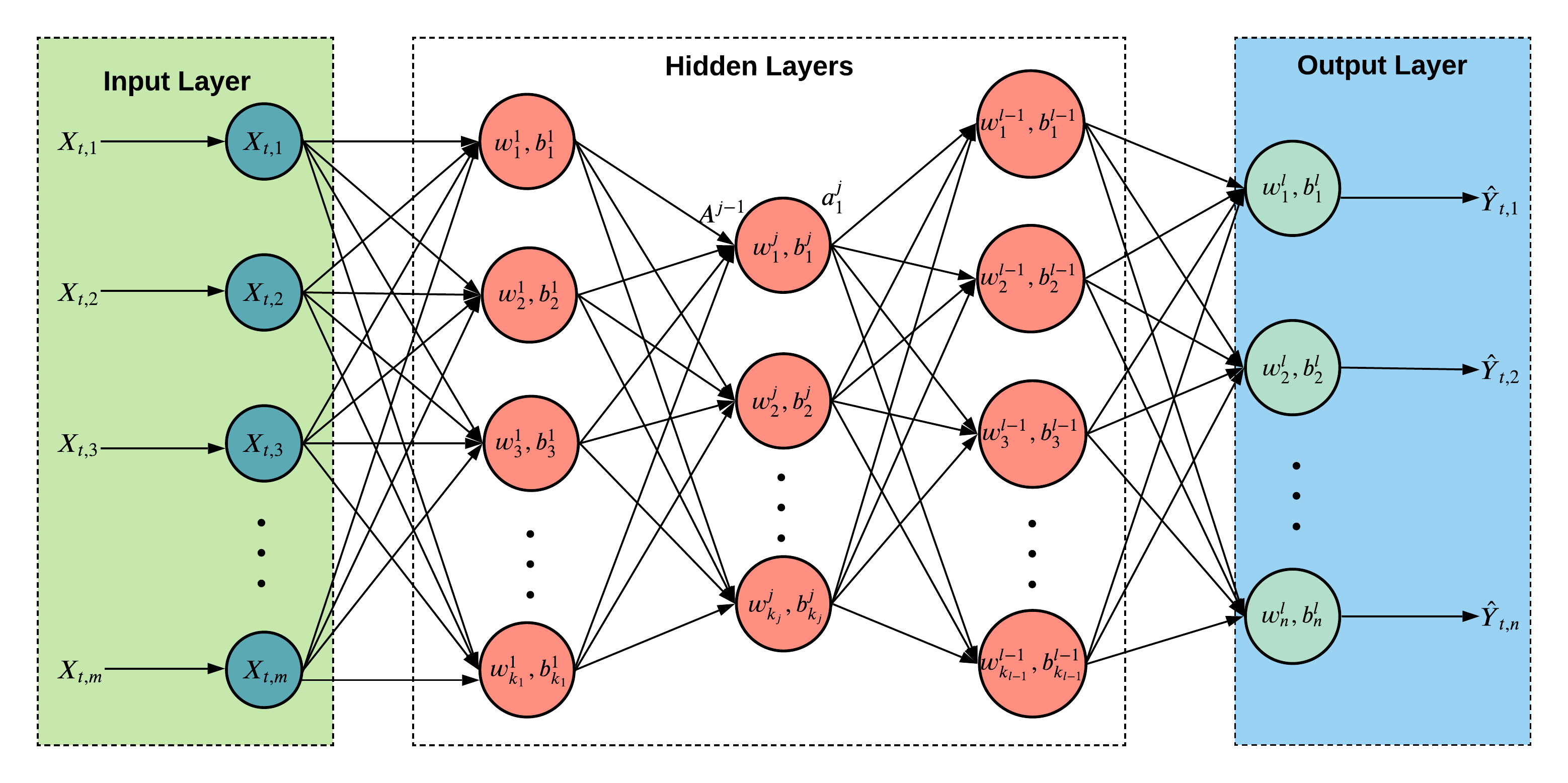}
		\caption{Feed-Forward Neural Network}
		\label{fig:ffnn_architecture}
	\end{center}
\end{figure*}

An FFNN can be described as the most basic type of an NN. The neurons within a single layer of an FFNN have no feedback connections as in an RNN and the neurons of each layer feed their outputs directly into the next immediate layer of the stack. 
Thus, FFNNs, when used for forecasting problems as described here, do not differentiate between individual time series as an RNN, since they do not have per-series states that propagate towards the end of the series. FFNNs can only distinguish between windows when used in a moving window scheme. The basic architecture of an FFNN is as illustrated in Figure \ref{fig:ffnn_architecture}. The FFNNs used for this study are implemented with version $2.0$ of the Tensorflow open-source deep learning framework~\citep{tensorflow2015-whitepaper}. 
%

\subsubsection{Light Gradient Boosting Models}

We also use LGBMs in our study as another forecasting technique. Extreme Gradient Boosting (XGBoost) models~\citep{xgboost2016} recently have become popular by winning many ML related challenges. Gradient Boosting Models (GBM) are a type of ML algorithms based on decision trees and can be used to solve a wide range of ML problems including regression, classification, and others. 
Due to the non-linear nature of decision trees, which is the fundamental building block of LGBMs, they too are non-linear models. 

The competitive performance of LGBMs when used for forecasting has been most recently demonstrated at the M5 Forecasting Competition, where an LGBM-based solution won the first place followed by many other LGBM-based solutions securing top ranks~\citep{makridakism5}. However, similar to FFNNs, LGBMs by default have no notion of the individual series of a dataset, when used for forecasting. They are only aware of the windows and they do not remember anything beyond what they see immediately within this input window. In our study we use the LGBM implementation available in the Python package \texttt{lightgbm}~\citep{lightgbm2017}. 


\subsubsection{Linear Auto-Regressive Models}

We also use univariate AR models, both non-seasonal and seasonal as other forecasting techniques in the study. The modelling process of these AR models is the same as described by Equations \ref{eqn:ar_model} and \ref{eqn:seasonal_ar_model} in the format of DGPs. AR models are linear in the lags of the target series. To implement the AR and SAR models for forecasting, we use the \texttt{Arima} function of the \texttt{forecast} package in the R programming language by providing the required seasonal or non-seasonal orders~\citep{forecast_package}.


\subsubsection{Self-Exciting Threshold Auto-Regressive Models}

For the scenarios involving the datasets generated from the SETAR DGP, we also experiment with SETAR as a forecasting technique. The underlying modelling process of the SETAR forecast model is as described in Equation \ref{eqn:tar_model} under Section \ref{sec:setar_dgp} for the DGPs. Due to the inherent regime-switching nature, SETAR is a non-linear forecast model. In this study, to implement the SETAR models we use the \texttt{setar} function from the \texttt{tsDyn} package in the R programming language~\citep{Fabio2019tsdyn}.

\subsubsection{Pooled Regression Models}

We implement PR models similar to the work of \citet{pooled_regression} as another forecasting technique for the study. 
The PR model is effectively a global version of an AR model.
The term pooling indicates that the coefficients of such regression models are calculated by pooling across many series. However, compared with, e.g., RNNs, the pooled regression models can only capture linear relationships in the data. Also, similar to FFNNs, PR models also do not maintain a state per every time series in the dataset.  A PR model of order $p$ can be defined as in Equation \ref{eqn:pooled_ar_model}. This equation is the same as in Equation \ref{eqn:ar_model} for the non-seasonal AR model, except now the coefficients $\Theta$ are calculated by pooling across all the series in the dataset.

\begin{equation}
y_{t} = c + \Theta_{1}By_{t} + \Theta_{2}B^{2}y_{t} + \Theta_{3}B^{3}y_{t} + ... + \Theta_{p}B^{p}y_{t} + \epsilon_{t}
\label{eqn:pooled_ar_model}
\end{equation}

PR models in this context are used to distinguish potential gains of using RNNs as opposed to linear models as global models. The PR models may also assist in identifying accuracy gains from purely using cross-series information without any complex architectural additions from RNNs~\citep{hewamalage2019recurrent}. To implement the PR models, we use the \texttt{glm} function in the \texttt{stats} package of the R core libraries~\citep{r_language}. 


The summary of all the forecasting techniques used for the study along with their respective attributes as well as the corresponding software details are available in Table \ref{tab:time_series_forecasting_techniques}.

\begin{table*}
	\footnotesize
	\begin{center}
		\begin{tabular}{ccc}
			\toprule
			\textbf{Forecasting Technique} & \textbf{Software Details}& \textbf{Attributes}\\
			\hline
			\multirow{2}{*}{RNN} & \multirow{2}{*}{\texttt{Tensorflow} Framework~\citep{tensorflow2015-whitepaper}} & Non-linear Modelling\\&&Per Series State\\
			\hline
			FFNN&\texttt{Tensorflow} Framework~\citep{tensorflow2015-whitepaper} &Non-linear Modelling \\
			\hline
			LGBM &\texttt{lightgbm} Python Package~\citep{lightgbm2017}&Non-linear Modelling\\
			\hline
			\multirow{2}{*}{SETAR} & \texttt{tsDyn} R Package~\citep{Fabio2019tsdyn} & Non-linear Modelling\\ & (\texttt{setar} function)&\\
			\hline
			\multirow{2}{*}{PR} & \texttt{stats} R Package~\citep{r_language} & Linear Modelling\\ & (\texttt{glm} function)&\\
			\hline
			\multirow{2}{*}{AR} & \texttt{forecast} R Package~\citep{forecast_package} & Linear Modelling\\ & (\texttt{Arima} function)&\\
			\hline
		\end{tabular}
		\caption{Summary of Forecasting Techniques}
		\label{tab:time_series_forecasting_techniques}
	\end{center}
\end{table*}

\subsection{Data Preprocessing for Forecasting}

We apply a number of data preprocessing steps to the data corresponding to the different forecasting techniques. 1) We first perform a normalization of each series to avoid the varying scales of the series, especially when developing a global model. 2) Then, we perform a logarithmic transformation of the series in order to avoid non-stationarities in time series data such as exponential trends that machine learning models cannot handle properly. 3) Next, to produce a forecast horizon, we perform a moving window transformation of the data specifically for those models that can output windows; other models use either a recursive startegy or train one model per each step in the horizon. 4) Finally, to further reduce risks from remaining linear trends, a per-window normalization is applied to the data which is first moving window transformed. A detailed description of all the preprocessing steps can be found in Appendix C of the Online Appendix.

\subsection{Model Training \& Testing}
RNNs in this study use the  COntinuous COin Betting (COCOB) optimiser introduced by \citet{Orabona2017-qi} as the underlying learning algorithm where as FFNNs use the well-known Adam optimiser~\citep{Kingma2014-pa}. For more details related to model training such as the hyperparameter tuning, loss functions used as well as the validation setups, refer to Appendix B of the Online Appendix.
For the NNs, once the optimal hyperparameters are found, they are applied for training the NNs once again and testing on the test data. During testing, the NNs are trained using 10 Tensorflow graph seeds and then the final NN forecasts are ensembled by taking the median, as discussed by \citet{hewamalage2019recurrent}. This approach effectively addresses the parameter uncertainty associated with the NNs by initialising the networks 10 times independently. 


For all the other models, once the model training is completed, the trained models are then used to perform the forecasting on the intended forecast horizon. As mentioned before, even though AR, SETAR and PR models train using one-step-ahead forecasts, during testing an output window is formed by performing a recursive strategy. LGBMs achieve the same objective by having one model per each step in the horizon.

\subsection{Data Postprocessing}

Once the forecasts are obtained from each individual model, a sequence of postprocessing is done on the forecasts, to reverse the preprocessing steps that were carried out on the data beforehand. This is done in the following order: 1) Add the last input point trend value back into the corresponding output windows. 2) Transform the data by taking the exponential. 3) Deduct 1, in case the original data contains zero values. 4) Re-scale the forecasts by multiplying by the series means.
However, the exact postprocessing steps performed on the forecasts of each model, depend on which preprocessing was done in the beginning.  

\section{Experimental Setup}
\label{sec:experimental_setup}

This section details the experimental framework used in this study. To achieve significant results, for every scenario we generate 1000 datasets having the described characteristics and average the results over 1000 runs. 
In the rest of this section, we present the details of the datasets generation, error metrics used for the evaluation and the statistical tests conducted for the significance of the differences.

\subsection{Generation of Datasets}
\label{sec:generated_datasets}


The characteristics of different datasets generated from all DGPs according to the experimental scenarios explained in Section~\ref{sec:overview_of_methodology}, are shown in Table~\ref{tab:generated_datasets}. Here, the \textit{Min.\@ Length} and \textit{Max.\@ Length} columns refer to the range of the time series lengths we select when training our models. Similarly, \textit{Min.\@ No.\@ of Series} and \textit{Max.\@ No.\@ of Series} is the range of number of time series we use in our experiments.

\begin{table*}[htp]
	\hspace*{-0.5cm}
	\scriptsize
	\begin{tabular}{lcccccc}
		\hline
		Scenario     & No. of DGPs & Min. Length & Max. Length & Min. No. of Series & Max. No. of Series & Forecast Horizon \\ \hline
		\multicolumn{7}{c}{AR(3) DGP}                                                                                       \\ \hline
		SS           & 1           & 18          & 1800        & 1                  & 1                  & 3                \\
		MS-Hom-Short & 1           & 18          & 18          & 1                  & 100                & 3                \\
		MS-Hom-Long  & 1           & 18          & 180         & 100                & 100                & 3                \\
		MS-Het       & 100         & 18          & 180         & 100                & 100                & 3                \\ \hline
		\multicolumn{7}{c}{SAR(1) DGP}                                                                                      \\ \hline
		SS           & 1           & 24          & 2400        & 1                  & 1                  & 3                \\
		MS-Hom-Short & 1           & 24          & 24          & 1                  & 100                & 3                \\
		MS-Hom-Long  & 1           & 24          & 240         & 100                & 100                & 3                \\
		MS-Het       & 100         & 24          & 240         & 100                & 100                & 3                \\ \hline
		\multicolumn{7}{c}{Chaotic Logistic Map DGP}                                                                        \\ \hline
		SS           & 1           & 60          & 6000        & 1                  & 1                  & 12               \\
		MS-Hom-Short & 1           & 60          & 60          & 1                  & 100                & 12               \\
		MS-Hom-Long  & 1           & 60          & 600         & 100                & 100                & 12               \\
		MS-Het       & 100         & 60          & 600         & 100                & 100                & 12               \\ \hline
		\multicolumn{7}{c}{SETAR DGP}                                                                                       \\ \hline
		SS           & 1           & 6000        & 6000        & 1                  & 1                  & 12               \\
		MS-Hom-Short & 1           & 60          & 60          & 100                & 100                & 12               \\
		MS-Hom-Long  & 1           & 240         & 240         & 100                & 100                & 12               \\
		MS-Het       & 100         & 240         & 240         & 100                & 100                & 12               \\ \hline
		\multicolumn{7}{c}{Mackey-Glass Equation DGP}                                                                       \\ \hline
		SS           & 1           & 6000        & 6000        & 1                  & 1                  & 12               \\
		MS-Hom-Short & 1           & 60          & 60          & 100                & 100                & 12               \\
		MS-Hom-Long  & 1           & 240         & 240         & 100                & 100                & 12               \\
		MS-Het       & 100         & 240         & 240         & 100                & 100                & 12               \\ \hline
	\end{tabular}
	\caption{Characteristics of Generated Datasets from All DGPs}
	\label{tab:generated_datasets}
\end{table*}

Due to computational constraints arising from the variety of experiments designed, out of the five DGPs used in the study, we perform the data availability experiments only using three DGPs, namely AR(3), SAR(1) and Chaotic Logistic Map. For SETAR and Mackey-Glass Equation DGPs, we experiment using only one selected length and number of series as shown in Table \ref{tab:generated_datasets}. Here, the idea is that once the important conclusions are drawn about data availability in initial experiments, those conclusions can be used to fix a sufficient length and the number of series for the remaining experiments. Also, due to the computational constraints, we perform the Group Feature setup experiments only using the Chaotic Logistic Map DGP. 
In Table~\ref{tab:generated_datasets_group_feature}, are the characteristics of the datasets generated for this Group Feature setup of the Chaotic Logistic Map DGP.

We also drop the number of datatsets associated with each scenario to 100, to avoid heavy computational complexities of this study. This applies to the experiments on the two DGPs, Mackey-Glass Equation and SETAR and the Group Feature setup of the Chaotic Logistic Map DGP.

\begin{table*}
	\begin{center}
		\footnotesize
		\begin{tabular}{lc}
			\toprule
			Characteristic & Value\\
			\hline
			No.\@ of DGPs & 4\\
			Min.\@ Length & 600\\
			Max.\@ Length & 600\\
			Min.\@ No.\@ of Series & 100\\
			Max.\@ No.\@ of Series & 100\\
			Forecast Horizon & 12\\
			\bottomrule
		\end{tabular}
		\caption{Characteristics of Generated Datasets for the Group Feature Setup of the Chaotic Logistic Map DGP}
		\label{tab:generated_datasets_group_feature}
	\end{center}
\end{table*}




\subsection{Performance Measures}

The relative performance of the models is evaluated with respect to two performance measures commonly found in the literature related to forecasting, namely SMAPE and Mean Absolute Scaled Error (MASE). Further details of the error measures can be found under Appendix D of the Online Appendix. However, According to, e.g., \citet{HYNDMAN2006ErrorMeasures}, the SMAPE metric is highly skewed with values close to zero. Therefore, as per the guidelines by \citet{hewamalage2019recurrent}, on the Chaotic Logistic Map DGP datasets which have zero values, we use a variant of SMAPE to address this problem. Since our forecasting scenarios involve multiple series from many datasets, we consider the mean SMAPE and mean MASE measures to summarise the overall error distribution across all the series available.

\subsection{Statistical Tests for Significance of the Differences}

The objective behind performing every experiment on 100/1000 datasets is that the individual differences between the models are significant for every scenario that way.
However, to show systematically that this is the case, we perform non-parametric Friedman rank-sum tests on few of the most important scenarios as explained in Section \ref{sec:results_analysis}, to estimate the significance of differences. This is followed by Hochberg's post hoc procedure to further analyse the differences relative to a control method, in particular the best method in every scenario~\citep{GARCIA20102044}. A significance level of $\alpha=0.05$ is used for all the tests.


\section{Results and Analysis}
\label{sec:results_analysis}

This section provides a detailed analysis of the results obtained for different experimental setups. Furthermore, we provide a comparison of the computational complexities of various forecasting variants employed in this study. 

\subsection{Comparison of Model Performance}

The results of all the DGPs as well as the percentage difference of every model from the best model in each scenario computed according to Equation \ref{eqn:percentage_difference_of_model_performance} are presented in Table \ref{tab:all_dgp_results_and_differences} under the \textit{Error} and \textit{Diff} columns. 
In Equation \ref{eqn:percentage_difference_of_model_performance}, $D_m$ is the percentage difference of the performance of model m, $E_m$ is the error of model $m$ and $E_b$ is the error of the best model in the respective scenario. 

\begin{equation}
	D_{m} = 100 * \frac{E_m - E_b}{E_b}
	\label{eqn:percentage_difference_of_model_performance}
\end{equation}

\begin{sidewaystable}[]
	\footnotesize
	\begin{tabular}{|l|cc|cc|cc|cc|cc|cc|cc|cc|}
		\hline
		Model    & \multicolumn{4}{c|}{SS}                                        & \multicolumn{4}{c|}{MS-Hom-Short}                              & \multicolumn{4}{c|}{MS-Hom-Long}                               & \multicolumn{4}{c|}{MS-Het}                                    \\ \cline{2-17} 
		& \multicolumn{2}{c|}{SMAPE}     & \multicolumn{2}{c|}{MASE}     & \multicolumn{2}{c|}{SMAPE}     & \multicolumn{2}{c|}{MASE}     & \multicolumn{2}{c}{SMAPE}      & \multicolumn{2}{c|}{MASE}     & \multicolumn{2}{c}{SMAPE}      & \multicolumn{2}{c|}{MASE}     \\ \cline{2-17} 
		& Error          & Diff          & Error         & Diff          & Error          & Diff          & Error         & Diff          & Error          & Diff          & Error         & Diff          & Error          & Diff          & Error         & Diff          \\ \cline{2-17} 
		\multicolumn{17}{|c|}{AR(3) DGP}                                                                                                                                                                                                                                             \\ \hline
		RNN(3)   & 18.04          & 3.26          & 0.55          & 3.77          & 18.12          & 2.32          & 0.61          & 1.67          & 21.37          & 1.42          & 0.54          & 1.89          & 21.78          & 9.28          & 0.86          & 13.16         \\
		RNN(10)  & -              & -             & -             & -             & -              & -             & -             & -             & 21.22          & 0.71          & 0.53          & 0.00          & 20.88          & 4.77          & 0.81          & 6.58          \\
		FFNN(3)  & 19.11          & 9.39          & 0.58          & 9.43          & 18.91          & 6.78          & 0.64          & 6.67          & 23.06          & 9.44          & 0.58          & 9.43          & 23.35          & 17.16         & 0.91          & 19.74         \\
		FFNN(10) & -              & -             & -             & -             & -              & -             & -             & -             & 21.84          & 3.65          & 0.55          & 3.77          & 22.21          & 11.44         & 0.87          & 14.47         \\
		LGBM(3)  & 19.25          & 10.19         & 0.59          & 11.32         & 19.25          & 8.70          & 0.65          & 8.33          & 23.07          & 9.49          & 0.58          & 9.43          & 23.29          & 16.86         & 0.90          & 18.42         \\
		LGBM(10) & -              & -             & -             & -             & -              & -             & -             & -             & 21.72          & 3.08          & 0.55          & 3.77          & 21.85          & 9.63          & 0.85          & 11.84         \\
		PR(3)    & -              & -             & -             & -             & \textbf{17.71} & \textbf{0.00} & \textbf{0.60} & \textbf{0.00} & \textbf{21.07} & \textbf{0.00} & \textbf{0.53} & \textbf{0.00} & 21.72          & 8.98          & 0.86          & 13.16         \\
		PR(10)   & -              & -             & -             & -             & -              & -             & -             & -             & 21.08          & 0.05          & \textbf{0.53} & \textbf{0.00} & 21.72          & 8.98          & 0.86          & 13.16         \\
		AR(2)    & 17.48          & 0.06          & \textbf{0.53} & \textbf{0.00} & 18.79          & 6.10          & 0.63          & 5.00          & 21.17          & 0.47          & \textbf{0.53} & \textbf{0.00} & \textbf{19.93} & \textbf{0.00} & \textbf{0.76} & \textbf{0.00} \\
		AR(3)    & \textbf{17.47} & \textbf{0.00} & \textbf{0.53} & \textbf{0.00} & 19.53          & 10.28         & 0.66          & 10.00         & 21.22          & 0.71          & \textbf{0.53} & \textbf{0.00} & 19.95          & 0.10          & \textbf{0.76} & \textbf{0.00} \\
		AR(10)   & 17.55          & 0.46          & 0.54          & 1.89          & 32.92          & 85.88         & 1.03          & 71.67         & 21.63          & 2.66          & 0.54          & 1.89          & 20.34          & 2.06          & 0.78          & 2.63          \\ \hline
		\multicolumn{17}{|c|}{SAR(1) DGP}                                                                                                                                                                                                                                            \\ \hline
		RNN(12)  & 4.91           & 3.37          & 0.97          & 3.19          & 4.90           & 0.82          & 1.04          & 0.97          & 14.03          & 6.45          & 1.04          & 7.22          & 22.15          & 4.48          & 0.74          & 5.71          \\
		FFNN(12) & 4.94           & 4.00          & 0.98          & 4.26          & \textbf{4.86}  & \textbf{0.00} & \textbf{1.03} & \textbf{0.00} & 13.22          & 0.30          & \textbf{0.97} & \textbf{0.00} & 22.98          & 8.40          & 0.77          & 10.00         \\
		LGBM(12) & 5.00           & 5.26          & 0.99          & 5.32          & 5.19           & 6.79          & 1.10          & 6.80          & 13.29          & 0.83          & 0.98          & 1.03          & 22.84          & 7.74          & 0.76          & 8.57          \\
		PR(12)   & -              & -             & -             & -             & 4.93           & 1.44          & 1.05          & 1.94          & 13.19          & 0.08          & \textbf{0.97} & \textbf{0.00} & 22.02          & 3.87          & 0.74          & 5.71          \\
		AR(12)   & 4.77           & 0.42          & \textbf{0.94} & \textbf{0.00} & 7.69           & 58.23         & 1.59          & 54.37         & 13.49          & 2.35          & 0.99          & 2.06          & 21.65          & 2.12          & 0.72          & 2.86          \\
		AR(3)    & 8.86           & 86.53         & 1.76          & 87.23         & 8.88           & 82.72         & 1.90          & 84.47         & 23.40          & 77.54         & 1.79          & 84.54         & 22.11          & 4.29          & 0.74          & 5.71          \\
		SAR(1)   & \textbf{4.75}  & \textbf{0.00} & \textbf{0.94} & \textbf{0.00} & 4.90           & 0.82          & 1.04          & 0.97          & \textbf{13.18} & \textbf{0.00} & \textbf{0.97} & \textbf{0.00} & \textbf{21.20} & \textbf{0.00} & \textbf{0.70} & \textbf{0.00} \\ \hline
		\multicolumn{17}{|c|}{Chaotic Logistic Map DGP}                                                                                                                                                                                                                              \\ \hline
		RNN(15)  & 51.63          & 5.52          & 0.78          & 5.41          & \textbf{49.21} & \textbf{0.00} & \textbf{0.75} & \textbf{0.00} & 48.12          & 2.82          & 0.73          & 2.82          & 27.54          & 0.33          & \textbf{0.88} & \textbf{0.00} \\
		FFNN(15) & 53.27          & 8.87          & 0.81          & 9.46          & 50.15          & 1.91          & 0.77          & 2.67          & 47.50          & 1.50          & 0.72          & 1.41          & 28.18          & 2.66          & 0.90          & 2.27          \\
		LGBM(15) & \textbf{48.93} & \textbf{0.00} & \textbf{0.74} & \textbf{0.00} & 50.43          & 2.48          & 0.77          & 2.67          & \textbf{46.80} & \textbf{0.00} & \textbf{0.71} & \textbf{0.00} & \textbf{27.45} & \textbf{0.00} & \textbf{0.88} & \textbf{0.00} \\
		PR(15)   & -              & -             & -             & -             & 53.37          & 8.45          & 0.81          & 8.00          & 52.46          & 12.09         & 0.78          & 9.86          & 27.63          & 0.66          & \textbf{0.88} & \textbf{0.00} \\
		AR(15)   & 52.91          & 8.13          & 0.79          & 6.76          & 57.67          & 17.19         & 0.89          & 18.67         & 52.65          & 12.50         & 0.79          & 11.27         & 27.56          & 0.40          & \textbf{0.88} & \textbf{0.00} \\ \hline
		\multicolumn{17}{|c|}{SETAR DGP}                                                                                                                                                                                                                                             \\ \hline
		RNN(15)  & 23.44          & 14.23         & 0.47          & 14.63         & \textbf{20.57} & \textbf{0.00} & \textbf{0.41} & \textbf{0.00} & \textbf{23.68} & \textbf{0.00} & \textbf{0.41} & \textbf{0.00} & 25.93          & 0.43          & \textbf{0.45} & \textbf{0.00} \\
		FFNN(15) & 20.78          & 1.27          & 0.42          & 2.44          & 20.69          & 0.58          & 0.42          & 2.44          & 23.84          & 0.68          & 0.42          & 2.44          & 26.00          & 0.70          & 0.46          & 2.22          \\
		LGBM(15) & 20.63          & 0.54          & \textbf{0.41} & \textbf{0.00} & 21.09          & 2.53          & 0.43          & 4.88          & 23.80          & 0.51          & 0.42          & 2.44          & \textbf{25.82} & \textbf{0.00} & \textbf{0.45} & \textbf{0.00} \\
		PR(15)   & -              & -             & -             & -             & 21.21          & 3.11          & 0.43          & 4.88          & 24.50          & 3.46          & 0.43          & 4.88          & 26.67          & 3.29          & 0.47          & 4.44          \\
		AR(15)   & 21.19          & 3.27          & 0.43          & 4.88          & 23.89          & 16.14         & 0.48          & 17.07         & 25.10          & 6.00          & 0.44          & 7.32          & 27.16          & 5.19          & 0.48          & 6.67          \\
		SETAR    & \textbf{20.52} & \textbf{0.00} & \textbf{0.41} & \textbf{0.00} & 22.26          & 8.22          & 0.44          & 7.32          & 25.61          & 8.15          & 0.44          & 7.32          & 29.10          & 12.70         & 0.51          & 13.33         \\ \hline
		\multicolumn{17}{|c|}{Mackey-Glass Equation DGP}                                                                                                                                                                                                                             \\ \hline
		RNN(15)  & 3.21           & 444.07        & 0.48          & 433.33        & 1.47           & 65.17         & 0.22          & 69.23         & 1.20           & 166.67        & 0.18          & 157.14        & \textbf{6.76}  & \textbf{0.00} & \textbf{1.11} & \textbf{0.00} \\
		FFNN(15) & 5.45           & 823.73        & 0.82          & 811.11        & 1.61           & 80.90         & 0.24          & 84.62         & 2.59           & 475.56        & 0.39          & 457.14        & 14.05          & 107.84        & 2.35          & 111.71        \\
		LGBM(15) & \textbf{0.59}  & \textbf{0.00} & \textbf{0.09} & \textbf{0.00} & \textbf{0.89}  & \textbf{0.00} & \textbf{0.13} & \textbf{0.00} & \textbf{0.45}  & \textbf{0.00} & \textbf{0.07} & \textbf{0.00} & 9.59           & 41.86         & 1.59          & 43.24         \\
		PR(15)   & -              & -             & -             & -             & 7.13           & 701.12        & 1.06          & 715.38        & 7.77           & 1626.67       & 1.16          & 1557.14       & 11.95          & 76.78         & 1.97          & 77.48         \\
		AR(15)   & 6.81           & 1054.24       & 1.02          & 1033.33       & 9.22           & 935.96        & 1.36          & 946.15        & 7.99           & 1675.56       & 1.19          & 1600.00       & 10.87          & 60.80         & 1.82          & 63.96         \\ \hline
	\end{tabular}
\caption{Results and Percentage Differences from the Best Model under each Experimental Scenario}
	\label{tab:all_dgp_results_and_differences}
\end{sidewaystable}

In every column of Table \ref{tab:all_dgp_results_and_differences}, the best-performing model under every DGP is indicated in boldface. Not all the models are relevant to all the experimental setups; hence the `-' in some of the table cells. Since PR models are the global versions of AR models, we run only the AR models in the SS scenarios. 
For the relevant models, we also indicate the number of lags used to train the model. For example, LGBM(3) refers to an LGBM model with an order of 3 lags. Although we conduct experiments for data availability for some of the DGPs, the numbers shown in Table \ref{tab:all_dgp_results_and_differences} correspond only to the highest length/number of series in each scenario as per Table \ref{tab:generated_datasets}. For more results analysis details, data availability experiment results as well as further details on the Hochberg's post-hoc procedures of different scenarios, refer to Appendix F of the Online Appendix.

\subsubsection{AR(3) Data Generating Process}

\begin{figure*}[htbp]
	\centering
	\captionsetup{justification=centering}
	\includegraphics[scale=0.6]{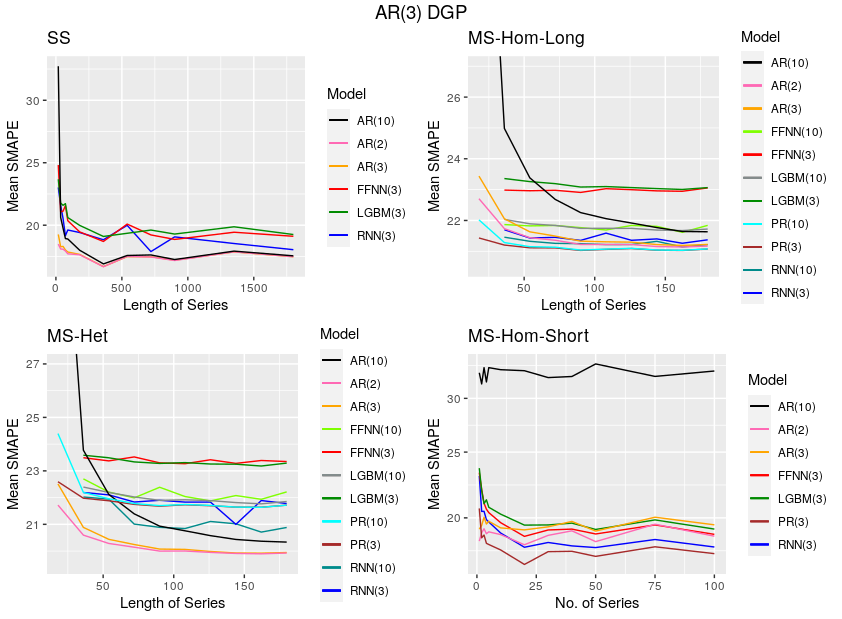}
	\caption{Visualisation of the Change of Errors of the Models Across Different Amounts of Data in the AR(3) DGP Scenarios}
	\label{fig:ar3_line_plot}
\end{figure*}

Figure \ref{fig:ar3_line_plot} illustrates how the performance of the forecasting models under the AR(3) DGP setting evolve with increasing lengths and number of series.

In Table \ref{tab:all_dgp_results_and_differences}, under the SS scenario of the AR(3) DGP, the AR(3) model performs the best, with the AR(2) being the second and the AR(10) the third, with respect to mean SMAPE. This is not surprising as AR(3) is the model of the true underlying DGP. Since AR(3) DGP is a subclass of the AR(10) forecast model, the training process of the AR(10) model resolves to training the coefficients beyond the third lag close to 0. 
However, when looking at the SS scenario in Figure \ref{fig:ar3_line_plot}, we can see that at the beginning lengths, the AR(2) model is better than the AR(3). 
For the SS scenario of the AR(3) DGP, the Friedman test gives an overall $p$-value of \num{1.21e-10} in terms of the mean SMAPE values indicating the significance of the differences of the models. Refer to Appendix F.3 for details of the Hochberg's post-hoc procedure results.

In the MS-Hom-Short case, the local AR model which learns only from the current series, becomes worse as the lengths of the individual series are much shorter (18 data points). Among the AR models as well, we can see how the performance gets worse with the increased number of coefficients; AR(10) is the worst and AR(3), the second worst. The PR(3) model performs best since the underlying DGP is an AR(3), so that a linear model of the corresponding order performs competitively. 
The RNN(3) model is the next best to the PR(3) in this setting. 
Furthermore, in a very short length scenario like this, it is evident how the global models such as PR(3) and RNN(3) which learn from all the series can be better than the local models, even AR(3) which is the one closest to the true DGP.
When the length of all the series is 18, AR(2) is better than AR(3) across all the dataset sizes, complying with the findings in the SS setting. In fact, at the beginning lengths, the AR(2) is the best model and gradually the PR(3) starts to outperform as the number of series in the dataset increases. AR(10) remains the worst performing model across all dataset sizes.

Since we have sufficient lengths in the MS-Hom-Long scenario, we run the PR, RNN, LGBM and FFNN models with both 3 and 10 lags. As seen in Table \ref{tab:all_dgp_results_and_differences}, the difference of the performance between the two models PR(3) and PR(10) is quite small in the MS-Hom-Long case. Similar to the MS-Hom-Short scenario, the PR models still remain the best out of all the models, in terms of Mean SMAPE. 
However, AR(2) and AR(3) models in the MS-Hom-Long scenario are much closer in performance to the PR models due to the increased lengths of the series (180 points). Yet, the PR model outperforms in terms of Mean SMAPE since the series are all homogeneous and the global, linear PR model learns better across the series. The lag 10 models all outperform their lag 3 variants for the complex models. In fact, with respect to mean MASE, RNN(10) demonstrates an equivalent performance to that of the best models PR(3), PR(10). In terms of Mean SMAPE, the RNN(10) model demonstrates the same result as the AR(3) model which is the true DGP. According to the MS-Hom-Long setting in Figure \ref{fig:ar3_line_plot}, we can see that AR(10) is the worst performing model in the beginning lengths similar to the observations from the MS-Hom-Short scenario. 
%

In the MS-Het scenario, the local AR(2) and AR(3) models again emerge as the best models with the AR(10) being the third. This can be explained by the induced heterogeneity among the series so that the local AR models built per every series outperform all the global models. Among the global models, the RNN(10) model performs better than the linear PR models since the inherent complex, non-linear modelling capacity of the RNN model can account for the heterogeneity among the series to a certain extent. According to the \textit{Diff} column in Table \ref{tab:all_dgp_results_and_differences}, the percentage difference between the best performing local models and the RNN(10) is very small (4.77\% in terms of Mean SMAPE) compared to the differences of other models. 
In Figure \ref{fig:ar3_line_plot}, although AR(10) outperforms all the global models at the maximum length, it performs the worst in the beginning lengths. Moreover, the PR(10) performs better than its corresponding local model AR(10) at the beginning lengths. Similarly, both RNN models perform better than AR(10) at shorter lengths. This demonstrates that although local models can be competitive in a heterogeneous series setting, for that they require sufficient lengths in the series. Otherwise, having global models, especially complex non-linear ones such as RNNs in a heterogeneous setting like this, which learn across all the short series together can be quite competitive. Friedman test for the MS-Het scenario gives an overall $p$-value of $\num{2.59e-10}$ in terms of the mean SMAPE values which means that the differences are statistically highly significant. For the details of Hochberg's post-hoc procedure, refer to Appendix F.3.




According to Figure \ref{fig:ar3_line_plot}, 
it is evident that the accuracies of all the models in general improve by increasing the length of the time series or the number of time series, except for the local models in the MS-Hom-Short scenario where we only increase the number of series in the dataset.


%
%


\subsubsection{SAR(1) Data Generating Process}

As seen in Table \ref{tab:all_dgp_results_and_differences} for the SAR(1) DGP, the SAR(1) model is the best on the SS scenario, which is obvious given that SAR(1) is the true DGP. The AR(3) model which is a misspecified model with respect to the SAR(1), is the worst. The AR(12) in the SS setting is equivalent in performance to SAR(1) in terms of Mean MASE. 
AR(12) is a superclass of the true DGP SAR(1), where only the 12th coefficient needs to be significant, which can be learned with sufficient length in the series. In the MS-Hom-Short scenario with the same data, the AR(3) model remains the worst performing model. However, the AR(12) has also become the second worst performing model in the MS-Hom-Short setting, due to the heavy complexity of the AR(12) model compared to the SAR(1) model and the short lengths of the series (24 points). The model closest to the true DGP which is SAR(1) shows the same performance as the RNN(12) and is the second best model. The FFNN outperforms the SAR(1) model which is the true DGP, although the short lengths of the series in the MS-Hom-Short setting are adequate to learn the one coefficient in the SAR(1) model. This shows the competence of the models that learn across series in such homogeneous contexts. The PR(12) model in the MS-Hom-Short scenario closely follows the SAR(1) and RNN(12) models.  

When the lengths of the series are increased as per the results of the MS-Hom-Long scenario, once again the SAR(1) model surpasses all the other models. However, the PR(12) model is almost the same as the SAR(1) model with the FFNN being the next following the PR(12). All three of them are equivalent in performance with respect to the Mean MASE values. 
In the MS-Het scenario, the SAR(1) model again becomes the best, complying with the observation in the AR(3) DGP. 
The second best model in this case is the AR(12) model, although it shows poor performance on the previously described homogeneous series scenarios. 
Among the global models, the PR(12) is the best. The RNN(12) is quite close and is exactly equal in performance to PR(12) with respect to Mean MASE. This indicates that, although the RNN is a competitive model due to its complexity in the heterogeneous setting, under the SAR(1) DGP with just one coefficient for every series, the RNN's high complexity may be unnecessary. Surprisingly, the misspecified AR(3) model outperforms all the non-linear global models in the MS-Het scenario. Theoretically this is unlikely for AR(3) even with the heterogeneous series.
However, this result is due to generating series in the MS-Het scenario by randomly picking a coefficient from U(-0.5, 0.5) to induce heterogeneity. Some of these series may end up getting non-significant 12th lags depending on how big the coefficient is. 
On the other hand in the MS-Hom-Long setting, the underlying SAR(1) model used has a coefficient of $0.85$ as mentioned in Appendix A.2.2 of our Online Appendix, which is chosen intentionally to create series with longer memory, and thus the inferior performance of AR(3) in that scenario. 
%
In the SAR(1) DGP too, the observations related to data availability are the same as for the AR(3) DGP, i.e., the performance of all the models improves in general as the size of the dataset increases, although not quite consistent across the individual lengths and number of series in some scenarios. 


\subsubsection{Chaotic Logistic Map Data Generating Process}

As shown in Table \ref{tab:all_dgp_results_and_differences}, the non-linearity of the patterns generated by the Chaotic Logistic Map DGP are perceived by the non-linear forecasting models outperforming the linear ones in all the different experimental scenarios. For the SS scenario, even though there is only one DGP with the same coefficients, the complex patterns produced by the DGP are not captured well by the linear model AR(15). The LGBM(15) is the best model on the SS scenario. On the MS-Hom-Short scenario, the RNN(15) is the best model with the FFNN(15) and the LGBM(15) being quite close. Once again, the lengths of the short series in the MS-Hom-Short scenario are not sufficient for the AR(15) model to properly learn all its 15 coefficients. However, the PR(15) model, although linear in nature, performs better than the AR(15) model in the MS-Hom-Short scenario, due to the underlying homegeneity of the series. Yet, since the individual series hold complex patterns, the linear PR is still not sufficient to surpass the complex global models in this scenario. For the MS-Hom-Short scenario, the Friedman test of statistical significance gives an overall $p$-value of \num{3.12e-10} indicating the significance of the differences between the models. Results of Hochberg's post-hoc procedure for this scenario indicate that all the models are significantly worse than the best model RNN(15) in this case. 

In the MS-Hom-Long case, still the non-linear complex models win over the linear models, while the LGBM(15) is the best. The drop of the errors in the AR model due to the length increase is notable. Despite their complexity, the lengths of the individual series in this case seem sufficient for the AR(15) to learn competitively. Yet, due to the homogeneity of the multiple series, the PR(15) which learns across series is ahead of the AR(15). For this scenario, the Friedman test of statistical significance gives an overall $p$-value of $<10^{-30}$. 
Hochberg's post-hoc procedure for this scenario, by using the LGBM(15) model as the control method indicate that all the other models are significantly worse than that. 
In the MS-Het scenario, even though the patterns become relatively easy to be modelled by all the forecasting models with respect to the mean SMAPE values (further explained in Appendix F.1.3), still the non-linear global models outperform the two linear AR and PR models. 
Furthermore, similar to the previous DGPs, the local AR(15) built per each series outperforms the PR(15). 

We can observe that the models improve with increased amounts of data in general, in the Chaotic Logistic Map DGP too.
Based on the conclusions obtained regarding data availability for forecasting from all the experimental scenarios explained thus far, we eliminate the data availability experiments from the rest of the experimental setups explained next. 

\subsubsection{SETAR Data Generating Process}


For the SS scenario of the SETAR DGP, the length of the series used is 6000. The best model under the SS scenario is the true model which is SETAR. 
Once again, the complexity of the DGP is visible by the non-linear models FFNN(15) and LGBM(15) outperforming the linear AR(15) model. However, the RNN(15) becomes the worst model in this scenario (further explained in Appendix F.1.4). Apart from the SS scenario, the SETAR DGP results are quite similar to the Chaotic Logistic Map DGP. For the MS-Hom-Short experiment, the length of every individual series is 60 with 100 series in the dataset (corresponding to the 6000 data points in the SS scenario). In this case, the RNN(15) is the best model followed by the FFNN(15) and the LGBM(15). The errors of the local AR and SETAR models increase due to short lengths of the series in the MS-Hom-Short scenario. Although the global PR is better than the local models due to the homogeneity of the series, it is not adequate to capture the complex patterns produced by the SETAR DGP for the individual series.

As seen in Table~\ref{tab:all_dgp_results_and_differences}, the non-linear global models surpass the linear AR and PR models in both the MS-Hom-Long and MS-Het scenarios. 
For the SETAR model the increased lengths in the MS-Hom-Long scenario still do not seem to be sufficient to capture the non-linear patterns of the series properly. The AR(15) model seems to learn some patterns better than at least the SETAR model with the increased lengths in the MS-Hom-Long scenario. Yet, they are both inferior in performance to the PR model which learns across series in this homogeneous case. However, in the heterogeneous scenario, different from the previous DGPs, the PR model performs better than both the AR and SETAR models. For the AR models, this can be explained by the regime-switching nature of the series generated by the SETAR DGP,
which makes one AR model trained for the whole series not the ideal model for this scenario.
Although the true model is the best under the heterogeneous settings of the previous DGPs, it is not the case with the SETAR DGP. The SETAR model in the heterogeneous case is still the worst performing model due to the insufficient lengths of the series. Also, it seems that the MS-Het scenario has made it difficult for all forecasting techniques to model the series in general.

\subsubsection{Mackey-Glass Data Generating Process}
\label{sec:mackey_glass_results}

The number and the lengths of the series in all the scenarios of the Mackey-Glass DGP are the same as in the SETAR DGP. The observations from this DGP too are very similar to SETAR and Chaotic Logistic Map DGPs. 
The non-linear models RNN(15), FFNN(15), and LGBM(15) outperform the linear models in most of the scenarios. The LGBM(15) model is remarkably better than all the other models under the SS, MS-Hom-Short and MS-Hom-Long scenarios. This is also confirmed by the Friedman test of statistical significance for the SS scenario, which gives an overall $p$-value of \num{1.08e-10}, suggesting that the results are statistically highly significant. Further tests with Hochberg's post-hoc procedure by using LGBM(15) as the control method indicate that all the other models are significantly worse than the LGBM(15) model in this case. 

The PR model is better than the AR model in both the MS-Hom-Short and MS-Hom-Long scenarios, due to the homogeneity among the series. 
The complexity of the datasets in the MS-Het scenario is evident by the substantial increase of the SMAPE values in all the models. In the heterogeneous scenario of the Mackey-Glass DGP too, the local AR model appears to be better than the global PR model, due to the inherent heterogeneity of the datasets. For the heterogeneous case, the Friedman test gives an overall $p$-value of \num{1.74e-10}, once again indicating the high significance of the performance differences. Hochberg's post-hoc process for the heterogeneous scenario reveals that all the other models are significantly worse than the RNN(15) model in this case.

\subsubsection{Results of the Group Feature Experiments}


The results from the Group Feature setup experiments performed on the Chaotic Logistic Map DGP are presented in Table \ref{tab:ms_het_few_results}. For this we use 100 datasets, each having series of length 240.
The horizontal line in Table \ref{tab:ms_het_few_results} separates the GFM.GroupFeature setup results with their corresponding GFM.All results along with the local AR model results. 
The ``GroupFeature" suffix indicates that the GFM.GroupFeature training paradigm explained in Section \ref{sec:complexity_of_forecasting_techniques} is used on the respective models.
As seen in Table \ref{tab:ms_het_few_results}, the GFM.GroupFeature setup improves the accuracy of the GFM.All setup in all techniques except the PR model where the SMAPE value remains constant. Especially in the case of the FFNN, the group feature inclusion results in the base model improving beyond the PR models. All the non-linear GroupFeature models hold a better standing over the linear models in this scenario, while the RNN(15)-Group performs the best. The Friedman test of statistical significance for the Group Feature scenario gives an overall $p$-value of \num{2.08e-10}, expressing that the differences are statistically highly significant. Refer to Appendix F.3 for further results with Hochberg's post-hoc procedure.



\begin{table}
	\begin{center}
		\footnotesize
		\begin{tabular}{l|cc}
			\toprule
			Model & SMAPE & MASE \\
			\hline
			RNN(15) & 38.27        & 0.87        \\
			FFNN(15) & 39.13        & 0.90        \\
			LGBM(15) & 37.39        & 0.86        \\
			PR(15) & 38.89        & 0.87        \\
			AR(15) & 39.31        & 0.88       \\
			\hline
			RNN(15)-Group & \textbf{36.68}        & \textbf{0.84}        \\
			FFNN(15)-Group & 38.40        & 0.88        \\
			LGBM(15)-Group & 37.05        & 0.85        \\
			PR(15)-Group & 38.89        & 0.87    \\
			\bottomrule   
		\end{tabular}
		\caption{Mean SMAPE and Mean MASE values for the GFM.GroupFeature Setup}
		\label{tab:ms_het_few_results}
	\end{center}

\end{table}

\subsection{Comparison of Computational Complexities}

Under the computational resources provided under Appendix E, we provide a comparison of the computational times of the different forecasting models. For this, we select the MS-Hom-Long scenario of the SETAR DGP.
In Table \ref{tab:computational_complexities}, we record the time taken for one of the 100 datasets.
Data preprocessing is not relevant for the AR(15) and SETAR models. Since RNN, FFNN, and LGBM all use the data preprocessed in the same manner, their data preprocessing times are the same. As observed from Table \ref{tab:computational_complexities}, the NN-based techniques spend comparatively higher computational times, with the RNN taking the highest and the FFNN the second highest time. 
The PR models in general are the most efficient, while the SETAR and the LGBM models are the second and third most efficient models respectively. Compared to LGBM and PR, which are global models, the local AR model in this case with its 15 coefficients also takes a considerable amount of time due to building one model per each series.


\begin{table*}
\footnotesize
	\begin{center}
		\begin{tabular}{lrrr}
			\toprule
			Model & Data Preprocessing & Model Training \& Testing & Total\\
			\hline
			RNN(15) & 5.36 & 10788.62 & 10793.98\\
			FFNN(15) & 5.36 & 649.73 & 655.09\\
			LGBM(15) & 5.36 & 4.58 & 9.94\\
			AR(15) & - & 111.08 & 111.08\\
			SETAR & - & 3.03 & 3.03 \\
			PR(15) &0.05& 0.40 & 0.45\\
			\hline
		\end{tabular}
		\caption{Computational Times Comparison of the Forecasting Models (in seconds)}
		\label{tab:computational_complexities}
	\end{center}
\end{table*}

\subsection{Overall Summary of Results}

In this section, we provide a summary for the overall results obtained. With respect to the data availability experiments, we can confirm the expected behaviour that the performance of all models, both local and global variants, improve as the individual series get longer. Also as expected, this is not the case with local models when we increase the number of series in the dataset, while keeping the lengths constant.
Considering the experiments related to single and multiple series, although the error values are not exactly the same, all global model accuracies are roughly equivalent in the SS and MS-Hom-Short scenario. This indicates that for global models it does not make much difference whether the data remains on one long series or spread across multiple different series. Several exceptions to this conclusion are also mentioned under Appendix F.2 of the Online Appendix.
The local AR models clearly become worse when moving to the MS-Hom-Short context, since less data is used for the model training then. Again, there are exceptions to this, such as the SAR(1) forecast model in the SAR(1) DGP. With its single coefficient, the SAR(1) model can be trained well with minimal training data. The lengths in the MS-Hom-Long settings can still be inadequate for certain local models depending on the underlying DGPs. An example of this is the MS-Hom-Long setting of the SETAR DGP where the SETAR model performs the worst.

From Table \ref{tab:all_dgp_results_and_differences}, we can see that with just a single series having adequate length, with either complex or simple patterns, the local forecast model closest to the true DGP if known, can win over all the other models. However, when moving into the MS-Hom-Short scenarios, this can happen only if the individual series are long enough. This depends on the lengths of the series as well as the complexity of the local model in terms of the number of coefficients. On the other hand, in multiple series scenarios, the power of global models which learn across series come into play. With these global models as well, the ones closest to the true underlying DGP can win.
The more complex the patterns in the individual series, the more complex global models can win, as seen with the MS-Hom-Short and MS-Hom-Long scenarios of the Chaotic Logistic Map DGP, SETAR DGP and Mackey-Glass Equation DGP. Even if the individual series lengths are sufficient for local models, if the series in the dataset are all homogeneous, the global models which learn across series are competitive over local ones which learn only from a single series. 
This holds for simple global models on series with complex patterns as well, as seen under the MS-Hom-Long scenarios of the Chaotic Logistic Map, SETAR and Mackey-Glass-Equation DGPs. When moving into the heterogeneous series settings, again the local models become competitive especially over the simple, linear global models. Once again, the potential of the global models come into play in the heterogeneous series settings, when the series lengths are short. However, unlike in the homogeneous settings, with the heterogeneous scenarios, the complex, non-linear global models are very competitive over simple, linear global models, with both simple and complex patterns in the individual series, as seen in the AR(3), Chaotic Logistic Map, SETAR and Mackey-Glass Equation DGPs. Apart from that, as seen in the AR(3) DGP case, for complex, non-linear global models, adding more memory/lags can help the models learn the patterns better.  
Adding external information to the GFM.All models regarding the existence of pre-known clusters within the data can further improve the model performance. This is confirmed by almost all the GFM.GroupFeature models outperforming their corresponding GFM.All setup models. 
Therefore, this attests our understanding that the GroupFeature setup can add more complexity to the base global model. 


Thus, the linear models such as PR, AR work best with known linear patterns in the data while non-linear models such as LGBM, RNN are more general in that they can perform reasonably well irrespective of the exact DGP. They can be competitive models under practical situations such as short series or heterogeneity across series. This is evident in Table \ref{tab:all_dgp_results_and_differences} where LGBMs and RNNs perform best under many scenarios. However, with RNNs, this superior performance comes at the cost of more computational time. LGBMs, on the other hand, are catching up with RNNs in terms of their performance, while being extremely fast in model training \& testing.

\section{Conclusions}
\label{sec:conclusions}

The recent work by \citet{monteromanso2020principles} has shown that any local method applied on a dataset of many series can be approximated by a global model with sufficient complexity, irrespective of the relatedness of the underlying series. Therefore, for global models it is about finding the right amount of complexity to outperform local methods. 
%
%
However, in practice there are complex trade-offs to be made between model complexity and capabilities on the one hand, and factors such as the availability of data, complexity of DGPs, and the heterogeneity of the underlying data on the other hand.
%
%
In this work, we have explored in an extensive experimental study some of the poorly understood factors that contribute to global forecasting model performance under these trade-offs. 
%
%
We have focused on characteristics common in real-world forecasting problems and their related challenges such as series with short history and heterogeneity of the series.
Through extensive empirical evaluations carried out within a controlled setting of simulated datasets, we have demonstrated the interplay between these different aspects.

In terms of the methodology, we start by simulating data using the arguably simplest DGPs available and then making them more complex. We start with simple, linear AR(3) and SAR(1) DGPs and then move onto more complex, non-linear DGPs with the Chaotic Logistic Map, SETAR, and Mackey-Glass Equations. 
We simulate both homogeneous and heterogeneous scenarios. In the homogeneous setup, all time series in the dataset are simulated using the same DGP, whereas in the heterogeneous case, time series from different DGPs are mixed together within the same dataset. The availability of data for the experiments is controlled by changing the lengths and the number of series. For each scenario, 100 or 1000 datasets are simulated using different random seeds, to achieve reliable and significant experimental results. Similar to the complexity of the DGPs, the complexity of the forecasting techniques is regulated by experimenting using a number of forecasting techniques with different modelling capabilities. We have used linear AR, PR models as well as more complex SETAR, RNNs, FFNNs, and LGBMs as forecasting techniques. The complexity of global models is further varied by introducing two model setups, GFM.All and GFM.GroupFeature. 

The results of our study have first confirmed that local linear models such as AR work best with known linear patterns in the data, with sufficient lengths in the series. The linear global models such as PR can cope well under multiple short series. When the patterns are made more complex, the complex non-linear global models such as RNNs, LGBMs are superior to simple linear global models. With heterogeneity existing across series, complex non-linear global models are competitive over linear global models irrespective of the simplicity or complexity of the patterns. Thus, non-linear, non-parametric models such as RNNs and LGBMs are in general quite competitive models across a variety of uncertain situations, where we have little prior knowledge about the data. The LGBM models hold another advantage of being computationally efficient compared to RNNs. Simple models such as AR and PR make assumptions about the linearity of the underlying data which are not always valid.
The results of the experiments that involve different scales of heterogeneity prove that the complexity of global forecasting models can be further improved by incorporating prior knowledge about the heterogeneity of data in the form of external features. With respect to data availability, unsurprisingly all global models gradually improve as the lengths and the number of series in the dataset increase. For local models, this improvement is understandably only seen with the length of the individual series. 
The model complexity of local models grows proportional to the number of series in the dataset, potentially even higher than the constant complexity of a global model built on the same dataset. This is why, even though fitting a single local model on just one series may take very little time, fitting many of them on a whole dataset of series takes a considerable amount of time. In terms of future work, further experiments can be conducted across all DGPs by changing the order of the same forecasting models within the same scenarios. For the heterogeneous setups, a third model setup GFM.Cluster can be introduced which trains multiple global models on the different clusters of the same dataset.






\section*{Acknowledgement}

This research was supported by the Australian Research Council under grant DE190100045, Facebook Statistics for Improving Insights and Decisions research award, Monash University Graduate Research funding and MASSIVE - High performance computing facility, Australia.

\begingroup
\setstretch{1.0}
\bibliographystyle{elsarticle-harv}
\bibliography{references}

\begin{thebibliography}{57}
\expandafter\ifx\csname natexlab\endcsname\relax\def\natexlab#1{#1}\fi
\providecommand{\url}[1]{\texttt{#1}}
\providecommand{\href}[2]{#2}
\providecommand{\path}[1]{#1}
\providecommand{\DOIprefix}{doi:}
\providecommand{\ArXivprefix}{arXiv:}
\providecommand{\URLprefix}{URL: }
\providecommand{\Pubmedprefix}{pmid:}
\providecommand{\doi}[1]{\href{http://dx.doi.org/#1}{\path{#1}}}
\providecommand{\Pubmed}[1]{\href{pmid:#1}{\path{#1}}}
\providecommand{\bibinfo}[2]{#2}
\ifx\xfnm\relax \def\xfnm[#1]{\unskip,\space#1}\fi
\bibitem[{Abadi et~al.(2015)Abadi, Agarwal, Barham, Brevdo, Chen, Citro,
  Corrado, Davis, Dean, Devin, Ghemawat, Goodfellow, Harp, Irving, Isard, Jia,
  Jozefowicz, Kaiser, Kudlur, Levenberg, Man\'{e}, Monga, Moore, Murray, Olah,
  Schuster, Shlens, Steiner, Sutskever, Talwar, Tucker, Vanhoucke, Vasudevan,
  Vi\'{e}gas, Vinyals, Warden, Wattenberg, Wicke, Yu and
  Zheng}]{tensorflow2015-whitepaper}
\bibinfo{author}{Abadi, M.}, \bibinfo{author}{Agarwal, A.},
  \bibinfo{author}{Barham, P.}, \bibinfo{author}{Brevdo, E.},
  \bibinfo{author}{Chen, Z.}, \bibinfo{author}{Citro, C.},
  \bibinfo{author}{Corrado, G.S.}, \bibinfo{author}{Davis, A.},
  \bibinfo{author}{Dean, J.}, \bibinfo{author}{Devin, M.},
  \bibinfo{author}{Ghemawat, S.}, \bibinfo{author}{Goodfellow, I.},
  \bibinfo{author}{Harp, A.}, \bibinfo{author}{Irving, G.},
  \bibinfo{author}{Isard, M.}, \bibinfo{author}{Jia, Y.},
  \bibinfo{author}{Jozefowicz, R.}, \bibinfo{author}{Kaiser, L.},
  \bibinfo{author}{Kudlur, M.}, \bibinfo{author}{Levenberg, J.},
  \bibinfo{author}{Man\'{e}, D.}, \bibinfo{author}{Monga, R.},
  \bibinfo{author}{Moore, S.}, \bibinfo{author}{Murray, D.},
  \bibinfo{author}{Olah, C.}, \bibinfo{author}{Schuster, M.},
  \bibinfo{author}{Shlens, J.}, \bibinfo{author}{Steiner, B.},
  \bibinfo{author}{Sutskever, I.}, \bibinfo{author}{Talwar, K.},
  \bibinfo{author}{Tucker, P.}, \bibinfo{author}{Vanhoucke, V.},
  \bibinfo{author}{Vasudevan, V.}, \bibinfo{author}{Vi\'{e}gas, F.},
  \bibinfo{author}{Vinyals, O.}, \bibinfo{author}{Warden, P.},
  \bibinfo{author}{Wattenberg, M.}, \bibinfo{author}{Wicke, M.},
  \bibinfo{author}{Yu, Y.}, \bibinfo{author}{Zheng, X.}, \bibinfo{year}{2015}.
\newblock \bibinfo{title}{{TensorFlow}: Large-scale machine learning on
  heterogeneous systems}.
\newblock \URLprefix \url{https://www.tensorflow.org/}. \bibinfo{note}{software
  available from tensorflow.org}.
\bibitem[{Bandara et~al.(2020a)Bandara, Bergmeir, Campbell, Scott and
  Lubman}]{Bandara2020-en}
\bibinfo{author}{Bandara, K.}, \bibinfo{author}{Bergmeir, C.},
  \bibinfo{author}{Campbell, S.}, \bibinfo{author}{Scott, D.},
  \bibinfo{author}{Lubman, D.}, \bibinfo{year}{2020}a.
\newblock \bibinfo{title}{Towards accurate predictions and causal 'what-if'
  analyses for planning and policy-making: A case study in emergency medical
  services demand}.
\bibitem[{Bandara et~al.(2020b)Bandara, Bergmeir and
  Smyl}]{bandara2020clustering}
\bibinfo{author}{Bandara, K.}, \bibinfo{author}{Bergmeir, C.},
  \bibinfo{author}{Smyl, S.}, \bibinfo{year}{2020}b.
\newblock \bibinfo{title}{Forecasting across time series databases using
  recurrent neural networks on groups of similar series: A clustering
  approach}.
\newblock \bibinfo{journal}{Expert Systems with Applications}
  \bibinfo{volume}{140}, \bibinfo{pages}{112896}.
\bibitem[{Bandara et~al.(2019)Bandara, Shi, Bergmeir, Hewamalage, Tran and
  Seaman}]{bandara2019ICONIP}
\bibinfo{author}{Bandara, K.}, \bibinfo{author}{Shi, P.},
  \bibinfo{author}{Bergmeir, C.}, \bibinfo{author}{Hewamalage, H.},
  \bibinfo{author}{Tran, Q.}, \bibinfo{author}{Seaman, B.},
  \bibinfo{year}{2019}.
\newblock \bibinfo{title}{Sales demand forecast in e-commerce using a long
  short-term memory neural network methodology}, in: \bibinfo{editor}{Gedeon,
  T.}, \bibinfo{editor}{Wong, K.W.}, \bibinfo{editor}{Lee, M.} (Eds.),
  \bibinfo{booktitle}{Neural Information Processing},
  \bibinfo{publisher}{Springer International Publishing},
  \bibinfo{address}{Cham}. pp. \bibinfo{pages}{462--474}.
\bibitem[{Bergmeir and Ben{\'i}tez(2012)}]{BERGMEIR2012CV}
\bibinfo{author}{Bergmeir, C.}, \bibinfo{author}{Ben{\'i}tez, J.M.},
  \bibinfo{year}{2012}.
\newblock \bibinfo{title}{On the use of cross-validation for time series
  predictor evaluation}.
\newblock \bibinfo{journal}{Information Sciences} \bibinfo{volume}{191},
  \bibinfo{pages}{192 -- 213}.
\newblock \bibinfo{note}{Data Mining for Software Trustworthiness}.
\bibitem[{Bergmeir et~al.(2018)Bergmeir, Hyndman and Koo}]{bergmeir2018cv}
\bibinfo{author}{Bergmeir, C.}, \bibinfo{author}{Hyndman, R.J.},
  \bibinfo{author}{Koo, B.}, \bibinfo{year}{2018}.
\newblock \bibinfo{title}{A note on the validity of cross-validation for
  evaluating autoregressive time series prediction}.
\newblock \bibinfo{journal}{Computational Statistics \& Data Analysis}
  \bibinfo{volume}{120}, \bibinfo{pages}{70 -- 83}.
\bibitem[{Bojer and Meldgaard(2020)}]{BOJER2020}
\bibinfo{author}{Bojer, C.S.}, \bibinfo{author}{Meldgaard, J.P.},
  \bibinfo{year}{2020}.
\newblock \bibinfo{title}{Kaggle forecasting competitions: An overlooked
  learning opportunity}.
\newblock \bibinfo{journal}{International Journal of Forecasting} .
\bibitem[{Box et~al.(1994)Box, Jenkins and Reinsel}]{Box1994-km}
\bibinfo{author}{Box, G.}, \bibinfo{author}{Jenkins, G.},
  \bibinfo{author}{Reinsel, G.}, \bibinfo{year}{1994}.
\newblock \bibinfo{title}{Time Series Analysis: Forecasting and Control}.
\newblock Forecasting and Control Series, \bibinfo{publisher}{Prentice Hall}.
\bibitem[{Chen and Guestrin(2016)}]{xgboost2016}
\bibinfo{author}{Chen, T.}, \bibinfo{author}{Guestrin, C.},
  \bibinfo{year}{2016}.
\newblock \bibinfo{title}{Xgboost: A scalable tree boosting system}, in:
  \bibinfo{booktitle}{Proceedings of the 22nd ACM SIGKDD International
  Conference on Knowledge Discovery and Data Mining},
  \bibinfo{publisher}{Association for Computing Machinery},
  \bibinfo{address}{New York, NY, USA}. p. \bibinfo{pages}{785–794}.
\newblock \DOIprefix\doi{10.1145/2939672.2939785}.
\bibitem[{Claveria and Torra(2014)}]{claveria2014}
\bibinfo{author}{Claveria, O.}, \bibinfo{author}{Torra, S.},
  \bibinfo{year}{2014}.
\newblock \bibinfo{title}{Forecasting tourism demand to catalonia: Neural
  networks vs. time series models}.
\newblock \bibinfo{journal}{Economic Modelling} \bibinfo{volume}{36},
  \bibinfo{pages}{220 -- 228}.
\bibitem[{Fabio Di~Narzo et~al.(2019)Fabio Di~Narzo, Luis~Aznarte and
  Stigler}]{Fabio2019tsdyn}
\bibinfo{author}{Fabio Di~Narzo, A.}, \bibinfo{author}{Luis~Aznarte, J.},
  \bibinfo{author}{Stigler, M.}, \bibinfo{year}{2019}.
\newblock \bibinfo{title}{tsDyn: Nonlinear Time Series Models with Regime
  Switching}.
\newblock \URLprefix \url{https://CRAN.R-project.org/package=tsDyn}.
  \bibinfo{note}{r package version 0.9-48.1}.
\bibitem[{Fischer et~al.(2018)Fischer, Krauss and
  Treichel}]{Fischer2018simulation}
\bibinfo{author}{Fischer, T.}, \bibinfo{author}{Krauss, C.},
  \bibinfo{author}{Treichel, A.}, \bibinfo{year}{2018}.
\newblock \bibinfo{title}{Machine learning for time series forecasting - a
  simulation study}.
\newblock \bibinfo{type}{FAU Discussion Papers in Economics}
  \bibinfo{number}{02/2018}. Friedrich-Alexander University Erlangen-Nuremberg,
  Institute for Economics.
\bibitem[{García et~al.(2010)García, Fernández, Luengo and
  Herrera}]{GARCIA20102044}
\bibinfo{author}{García, S.}, \bibinfo{author}{Fernández, A.},
  \bibinfo{author}{Luengo, J.}, \bibinfo{author}{Herrera, F.},
  \bibinfo{year}{2010}.
\newblock \bibinfo{title}{Advanced nonparametric tests for multiple comparisons
  in the design of experiments in computational intelligence and data mining:
  Experimental analysis of power}.
\newblock \bibinfo{journal}{Information Sciences} \bibinfo{volume}{180},
  \bibinfo{pages}{2044 -- 2064}.
\newblock \bibinfo{note}{Special Issue on Intelligent Distributed Information
  Systems}.
\bibitem[{Gers et~al.(2003)Gers, Schraudolph and
  Schmidhuber}]{Gers2003peephole}
\bibinfo{author}{Gers, F.A.}, \bibinfo{author}{Schraudolph, N.N.},
  \bibinfo{author}{Schmidhuber, J.}, \bibinfo{year}{2003}.
\newblock \bibinfo{title}{Learning precise timing with lstm recurrent
  networks}.
\newblock \bibinfo{journal}{J. Mach. Learn. Res.} \bibinfo{volume}{3},
  \bibinfo{pages}{115–143}.
\bibitem[{{He} et~al.(2016){He}, {Zhang}, {Ren} and {Sun}}]{He2016Resnet}
\bibinfo{author}{{He}, K.}, \bibinfo{author}{{Zhang}, X.},
  \bibinfo{author}{{Ren}, S.}, \bibinfo{author}{{Sun}, J.},
  \bibinfo{year}{2016}.
\newblock \bibinfo{title}{Deep residual learning for image recognition}, in:
  \bibinfo{booktitle}{2016 IEEE Conference on Computer Vision and Pattern
  Recognition (CVPR)}, pp. \bibinfo{pages}{770--778}.
\newblock \DOIprefix\doi{10.1109/CVPR.2016.90}.
\bibitem[{Hewamalage et~al.(2020)Hewamalage, Bergmeir and
  Bandara}]{hewamalage2019recurrent}
\bibinfo{author}{Hewamalage, H.}, \bibinfo{author}{Bergmeir, C.},
  \bibinfo{author}{Bandara, K.}, \bibinfo{year}{2020}.
\newblock \bibinfo{title}{Recurrent neural networks for time series
  forecasting: Current status and future directions}.
\newblock \bibinfo{journal}{International Journal of Forecasting} .
\bibitem[{Hyndman et~al.(2020)Hyndman, Athanasopoulos, Bergmeir, Caceres,
  Chhay, O'Hara-Wild, Petropoulos, Razbash, Wang and
  Yasmeen}]{forecast_package}
\bibinfo{author}{Hyndman, R.}, \bibinfo{author}{Athanasopoulos, G.},
  \bibinfo{author}{Bergmeir, C.}, \bibinfo{author}{Caceres, G.},
  \bibinfo{author}{Chhay, L.}, \bibinfo{author}{O'Hara-Wild, M.},
  \bibinfo{author}{Petropoulos, F.}, \bibinfo{author}{Razbash, S.},
  \bibinfo{author}{Wang, E.}, \bibinfo{author}{Yasmeen, F.},
  \bibinfo{year}{2020}.
\newblock \bibinfo{title}{{forecast}: Forecasting functions for time series and
  linear models}.
\newblock \URLprefix \url{http://pkg.robjhyndman.com/forecast}.
  \bibinfo{note}{r package version 8.11}.
\bibitem[{Hyndman et~al.(2008)Hyndman, Koehler, Ord and
  D~Snyder}]{hyndman2008ets}
\bibinfo{author}{Hyndman, R.}, \bibinfo{author}{Koehler, A.},
  \bibinfo{author}{Ord, K.}, \bibinfo{author}{D~Snyder, R.},
  \bibinfo{year}{2008}.
\newblock \bibinfo{title}{Forecasting with exponential smoothing. The state
  space approach}. \bibinfo{publisher}{Springer Berlin Heidelberg}.
\newblock \DOIprefix\doi{10.1007/978-3-540-71918-2}.
\bibitem[{Hyndman and Athanasopoulos(2018)}]{robgeorg2018otext}
\bibinfo{author}{Hyndman, R.J.}, \bibinfo{author}{Athanasopoulos, G.},
  \bibinfo{year}{2018}.
\newblock \bibinfo{title}{Forecasting: Principles and Practice}.
\newblock \bibinfo{edition}{second} ed., \bibinfo{publisher}{OTexts}.
\newblock \URLprefix \url{https://otexts.com/fpp2/}.
\bibitem[{Hyndman and Koehler(2006)}]{HYNDMAN2006ErrorMeasures}
\bibinfo{author}{Hyndman, R.J.}, \bibinfo{author}{Koehler, A.B.},
  \bibinfo{year}{2006}.
\newblock \bibinfo{title}{Another look at measures of forecast accuracy}.
\newblock \bibinfo{journal}{International Journal of Forecasting}
  \bibinfo{volume}{22}, \bibinfo{pages}{679 -- 688}.
\bibitem[{Hyndman et~al.(2002)Hyndman, Koehler, Snyder and
  Grose}]{hyndman2002ets}
\bibinfo{author}{Hyndman, R.J.}, \bibinfo{author}{Koehler, A.B.},
  \bibinfo{author}{Snyder, R.D.}, \bibinfo{author}{Grose, S.},
  \bibinfo{year}{2002}.
\newblock \bibinfo{title}{A state space framework for automatic forecasting
  using exponential smoothing methods}.
\newblock \bibinfo{journal}{International Journal of Forecasting}
  \bibinfo{volume}{18}, \bibinfo{pages}{439 -- 454}.
\bibitem[{Januschowski et~al.(2020)Januschowski, Gasthaus, Wang, Salinas,
  Flunkert, Bohlke-Schneider and Callot}]{Januschowski2020ijf}
\bibinfo{author}{Januschowski, T.}, \bibinfo{author}{Gasthaus, J.},
  \bibinfo{author}{Wang, Y.}, \bibinfo{author}{Salinas, D.},
  \bibinfo{author}{Flunkert, V.}, \bibinfo{author}{Bohlke-Schneider, M.},
  \bibinfo{author}{Callot, L.}, \bibinfo{year}{2020}.
\newblock \bibinfo{title}{Criteria for classifying forecasting methods}.
\newblock \bibinfo{journal}{International Journal of Forecasting}
  \bibinfo{volume}{36}, \bibinfo{pages}{167 -- 177}.
\newblock \bibinfo{note}{M4 Competition}.
\bibitem[{Kang et~al.(2020)Kang, Li and Hyndman}]{kang2019gratis}
\bibinfo{author}{Kang, Y.}, \bibinfo{author}{Li, F.}, \bibinfo{author}{Hyndman,
  R.J.}, \bibinfo{year}{2020}.
\newblock \bibinfo{title}{{GRATIS}: Generating time series with diverse and
  controllable characteristics}.
\newblock \bibinfo{journal}{Statistical Analysis and Data Mining}
  \bibinfo{volume}{13}, \bibinfo{pages}{354--376}.
\newblock \DOIprefix\doi{10.1002/sam.11461}.
\bibitem[{Ke et~al.(2017)Ke, Meng, Finley, Wang, Chen, Ma, Ye and
  Liu}]{lightgbm2017}
\bibinfo{author}{Ke, G.}, \bibinfo{author}{Meng, Q.}, \bibinfo{author}{Finley,
  T.}, \bibinfo{author}{Wang, T.}, \bibinfo{author}{Chen, W.},
  \bibinfo{author}{Ma, W.}, \bibinfo{author}{Ye, Q.}, \bibinfo{author}{Liu,
  T.Y.}, \bibinfo{year}{2017}.
\newblock \bibinfo{title}{Lightgbm: A highly efficient gradient boosting
  decision tree}, in: \bibinfo{editor}{Guyon, I.}, \bibinfo{editor}{Luxburg,
  U.V.}, \bibinfo{editor}{Bengio, S.}, \bibinfo{editor}{Wallach, H.},
  \bibinfo{editor}{Fergus, R.}, \bibinfo{editor}{Vishwanathan, S.},
  \bibinfo{editor}{Garnett, R.} (Eds.), \bibinfo{booktitle}{Advances in Neural
  Information Processing Systems 30}. \bibinfo{publisher}{Curran Associates,
  Inc.}, pp. \bibinfo{pages}{3146--3154}.
\bibitem[{Kingma and Ba(2015)}]{Kingma2014-pa}
\bibinfo{author}{Kingma, D.P.}, \bibinfo{author}{Ba, J.}, \bibinfo{year}{2015}.
\newblock \bibinfo{title}{Adam: A method for stochastic optimization}, in:
  \bibinfo{booktitle}{3rd International Conference for Learning
  Representations}.
\bibitem[{Lau and Wu(2008)}]{LAU20081539}
\bibinfo{author}{Lau, K.}, \bibinfo{author}{Wu, Q.}, \bibinfo{year}{2008}.
\newblock \bibinfo{title}{Local prediction of non-linear time series using
  support vector regression}.
\newblock \bibinfo{journal}{Pattern Recognition} \bibinfo{volume}{41},
  \bibinfo{pages}{1539 -- 1547}.
\bibitem[{Mackey and Glass(1977)}]{mackey1977}
\bibinfo{author}{Mackey, M.C.}, \bibinfo{author}{Glass, L.},
  \bibinfo{year}{1977}.
\newblock \bibinfo{title}{Oscillation and chaos in physiological control
  systems}.
\newblock \bibinfo{journal}{Science} \bibinfo{volume}{197},
  \bibinfo{pages}{287--289}.
\bibitem[{Makridakis et~al.(2020a)Makridakis, Spiliotis and
  Assimakopoulos}]{makridakis2020m4}
\bibinfo{author}{Makridakis, S.}, \bibinfo{author}{Spiliotis, E.},
  \bibinfo{author}{Assimakopoulos, V.}, \bibinfo{year}{2020}a.
\newblock \bibinfo{title}{The m4 competition: 100,000 time series and 61
  forecasting methods}.
\newblock \bibinfo{journal}{International Journal of Forecasting}
  \bibinfo{volume}{36}, \bibinfo{pages}{54 -- 74}.
\newblock \bibinfo{note}{M4 Competition}.
\bibitem[{Makridakis et~al.(2020b)Makridakis, Spiliotis and
  Assimakopoulos}]{makridakism5}
\bibinfo{author}{Makridakis, S.}, \bibinfo{author}{Spiliotis, E.},
  \bibinfo{author}{Assimakopoulos, V.}, \bibinfo{year}{2020}b.
\newblock \bibinfo{title}{The m5 accuracy competition: Results, findings and
  conclusions}.
\newblock \URLprefix
  \url{https://www.researchgate.net/publication/344487258_The_M5_Accuracy_competition_Results_findings_and_conclusions}.
\bibitem[{Mandal et~al.(2006)Mandal, Senjyu, Urasaki and
  Funabashi}]{Mandal2006}
\bibinfo{author}{Mandal, P.}, \bibinfo{author}{Senjyu, T.},
  \bibinfo{author}{Urasaki, N.}, \bibinfo{author}{Funabashi, T.},
  \bibinfo{year}{2006}.
\newblock \bibinfo{title}{A neural network based several-hour-ahead electric
  load forecasting using similar days approach}.
\newblock \bibinfo{journal}{International Journal of Electrical Power \& Energy
  Systems} \bibinfo{volume}{28}, \bibinfo{pages}{367 -- 373}.
\bibitem[{Mannattil(2017)}]{nolitsa}
\bibinfo{author}{Mannattil, M.}, \bibinfo{year}{2017}.
\newblock \bibinfo{title}{nolitsa}.
\newblock
  \bibinfo{howpublished}{\url{https://github.com/manu-mannattil/nolitsa}}.
\bibitem[{May(1976)}]{May1976}
\bibinfo{author}{May, R.M.}, \bibinfo{year}{1976}.
\newblock \bibinfo{title}{Simple mathematical models with very complicated
  dynamics}.
\newblock \bibinfo{journal}{Nature} \bibinfo{volume}{261},
  \bibinfo{pages}{459--467}.
\bibitem[{Montero-Manso and Hyndman(2020)}]{monteromanso2020principles}
\bibinfo{author}{Montero-Manso, P.}, \bibinfo{author}{Hyndman, R.J.},
  \bibinfo{year}{2020}.
\newblock \bibinfo{title}{Principles and algorithms for forecasting groups of
  time series: Locality and globality}.
\newblock \URLprefix \url{https://arxiv.org/abs/2008.00444},
  \href{http://arxiv.org/abs/2008.00444}{{\tt arXiv:2008.00444}}.
\bibitem[{Orabona and Tommasi(2017)}]{Orabona2017-qi}
\bibinfo{author}{Orabona, F.}, \bibinfo{author}{Tommasi, T.},
  \bibinfo{year}{2017}.
\newblock \bibinfo{title}{Training deep networks without learning rates through
  coin betting}, in: \bibinfo{booktitle}{Proceedings of the 31st International
  Conference on Neural Information Processing Systems},
  \bibinfo{publisher}{Curran Associates Inc.}, \bibinfo{address}{USA}. pp.
  \bibinfo{pages}{2157--2167}.
\bibitem[{Petropoulos et~al.(2014)Petropoulos, Makridakis, Assimakopoulos and
  Nikolopoulos}]{Petropoulos2014horses}
\bibinfo{author}{Petropoulos, F.}, \bibinfo{author}{Makridakis, S.},
  \bibinfo{author}{Assimakopoulos, V.}, \bibinfo{author}{Nikolopoulos, K.},
  \bibinfo{year}{2014}.
\newblock \bibinfo{title}{‘horses for courses’ in demand forecasting}.
\newblock \bibinfo{journal}{European Journal of Operational Research}
  \bibinfo{volume}{237}, \bibinfo{pages}{152 -- 163}.
\bibitem[{{R Core Team}(2020)}]{r_language}
\bibinfo{author}{{R Core Team}}, \bibinfo{year}{2020}.
\newblock \bibinfo{title}{R: A Language and Environment for Statistical
  Computing}.
\newblock \bibinfo{organization}{R Foundation for Statistical Computing}.
  \bibinfo{address}{Vienna, Austria}.
\newblock \URLprefix \url{https://www.R-project.org/}.
\bibitem[{Rangapuram et~al.(2018)Rangapuram, Seeger, Gasthaus, Stella, Wang and
  Januschowski}]{Rangapuram2018deep}
\bibinfo{author}{Rangapuram, S.S.}, \bibinfo{author}{Seeger, M.},
  \bibinfo{author}{Gasthaus, J.}, \bibinfo{author}{Stella, L.},
  \bibinfo{author}{Wang, Y.}, \bibinfo{author}{Januschowski, T.},
  \bibinfo{year}{2018}.
\newblock \bibinfo{title}{Deep state space models for time series forecasting},
  in: \bibinfo{booktitle}{Proceedings of the 32nd International Conference on
  Neural Information Processing Systems}, \bibinfo{publisher}{Curran Associates
  Inc.}, \bibinfo{address}{Red Hook, NY, USA}. p. \bibinfo{pages}{7796–7805}.
\bibitem[{Salinas et~al.(2019)Salinas, Flunkert, Gasthaus and
  Januschowski}]{Flunkert2017-wp}
\bibinfo{author}{Salinas, D.}, \bibinfo{author}{Flunkert, V.},
  \bibinfo{author}{Gasthaus, J.}, \bibinfo{author}{Januschowski, T.},
  \bibinfo{year}{2019}.
\newblock \bibinfo{title}{Deepar: Probabilistic forecasting with autoregressive
  recurrent networks}.
\newblock \bibinfo{journal}{International Journal of Forecasting} .
\bibitem[{Sch\"{a}fer and Zimmermann(2006)}]{Schafer2006-ln}
\bibinfo{author}{Sch\"{a}fer, A.M.}, \bibinfo{author}{Zimmermann, H.G.},
  \bibinfo{year}{2006}.
\newblock \bibinfo{title}{Recurrent neural networks are universal
  approximators}, in: \bibinfo{booktitle}{Proceedings of the 16th International
  Conference on Artificial Neural Networks - Volume Part I},
  \bibinfo{publisher}{Springer-Verlag}, \bibinfo{address}{Berlin, Heidelberg}.
  pp. \bibinfo{pages}{632--640}.
\newblock \DOIprefix\doi{10.1007/11840817_66}.
\bibitem[{Smyl(2020)}]{smyl2020esrnn}
\bibinfo{author}{Smyl, S.}, \bibinfo{year}{2020}.
\newblock \bibinfo{title}{A hybrid method of exponential smoothing and
  recurrent neural networks for time series forecasting}.
\newblock \bibinfo{journal}{International Journal of Forecasting}
  \bibinfo{volume}{36}, \bibinfo{pages}{75 -- 85}.
\newblock \bibinfo{note}{M4 Competition}.
\bibitem[{Smyl and Kuber(2016)}]{Smyl2016mcmc}
\bibinfo{author}{Smyl, S.}, \bibinfo{author}{Kuber, K.}, \bibinfo{year}{2016}.
\newblock \bibinfo{title}{Data preprocessing and augmentation for multiple
  short time series forecasting with recurrent neural networks}, in:
  \bibinfo{booktitle}{36th International Symposium on Forecasting}.
\bibitem[{Suhartono et~al.(2018)Suhartono, Amalia, Saputri, Rahayu and
  Ulama}]{Suhartono2018simulation}
\bibinfo{author}{Suhartono}, \bibinfo{author}{Amalia, F.F.},
  \bibinfo{author}{Saputri, P.D.}, \bibinfo{author}{Rahayu, S.P.},
  \bibinfo{author}{Ulama, B.S.S.}, \bibinfo{year}{2018}.
\newblock \bibinfo{title}{Simulation study for determining the best
  architecture of multilayer perceptron for forecasting nonlinear seasonal time
  series}.
\newblock \bibinfo{journal}{Journal of Physics: Conference Series}
  \bibinfo{volume}{1028}, \bibinfo{pages}{012214}.
\bibitem[{Sun et~al.(2019)Sun, Yang, Liu, Chen, Rao and Bai}]{SUN201924}
\bibinfo{author}{Sun, J.}, \bibinfo{author}{Yang, Y.}, \bibinfo{author}{Liu,
  Y.}, \bibinfo{author}{Chen, C.}, \bibinfo{author}{Rao, W.},
  \bibinfo{author}{Bai, Y.}, \bibinfo{year}{2019}.
\newblock \bibinfo{title}{Univariate time series classification using
  information geometry}.
\newblock \bibinfo{journal}{Pattern Recognition} \bibinfo{volume}{95},
  \bibinfo{pages}{24 -- 35}.
\bibitem[{Svetunkov(2019)}]{Ivan2019smooth}
\bibinfo{author}{Svetunkov, I.}, \bibinfo{year}{2019}.
\newblock \bibinfo{title}{smooth: Forecasting Using State Space Models}.
\newblock \URLprefix \url{https://CRAN.R-project.org/package=smooth}.
  \bibinfo{note}{r package version 2.5.3}.
\bibitem[{Tong(1978)}]{Tong1978TAR}
\bibinfo{author}{Tong, H.}, \bibinfo{year}{1978}.
\newblock \bibinfo{title}{On a threshold model}, in:
  \bibinfo{booktitle}{Pattern Recognition and Signal Processing}.
  \bibinfo{publisher}{Springer Netherlands}, pp. \bibinfo{pages}{575--586}.
\bibitem[{Tong and Lim(1980)}]{Tong1980SETAR}
\bibinfo{author}{Tong, H.}, \bibinfo{author}{Lim, K.S.}, \bibinfo{year}{1980}.
\newblock \bibinfo{title}{Threshold autoregression, limit cycles and cyclical
  data}.
\newblock \bibinfo{journal}{Journal of the Royal Statistical Society: Series B
  (Methodological)} \bibinfo{volume}{42}, \bibinfo{pages}{245--268}.
\bibitem[{Trapero et~al.(2015)Trapero, Kourentzes and
  Fildes}]{pooled_regression}
\bibinfo{author}{Trapero, J.R.}, \bibinfo{author}{Kourentzes, N.},
  \bibinfo{author}{Fildes, R.}, \bibinfo{year}{2015}.
\newblock \bibinfo{title}{On the identification of sales forecasting models in
  the presence of promotions}.
\newblock \bibinfo{journal}{Journal of the Operational Research Society}
  \bibinfo{volume}{66}, \bibinfo{pages}{299--307}.
\newblock \DOIprefix\doi{10.1057/jors.2013.174}.
\bibitem[{Vanli et~al.(2019)Vanli, Sayin, {Mohaghegh N.}, Ozkan and
  Kozat}]{VANLI20191}
\bibinfo{author}{Vanli, N.D.}, \bibinfo{author}{Sayin, M.O.},
  \bibinfo{author}{{Mohaghegh N.}, M.}, \bibinfo{author}{Ozkan, H.},
  \bibinfo{author}{Kozat, S.S.}, \bibinfo{year}{2019}.
\newblock \bibinfo{title}{Nonlinear regression via incremental decision trees}.
\newblock \bibinfo{journal}{Pattern Recognition} \bibinfo{volume}{86},
  \bibinfo{pages}{1 -- 13}.
\bibitem[{Wang et~al.(2019)Wang, Smola, Maddix, Gasthaus, Foster and
  Januschowski}]{pmlr-v97-wang19k}
\bibinfo{author}{Wang, Y.}, \bibinfo{author}{Smola, A.},
  \bibinfo{author}{Maddix, D.}, \bibinfo{author}{Gasthaus, J.},
  \bibinfo{author}{Foster, D.}, \bibinfo{author}{Januschowski, T.},
  \bibinfo{year}{2019}.
\newblock \bibinfo{title}{Deep factors for forecasting},
  \bibinfo{publisher}{PMLR}, \bibinfo{address}{Long Beach, California, USA}.
  pp. \bibinfo{pages}{6607--6617}.
\bibitem[{Wen et~al.(2017a)Wen, Torkkola, Narayanaswamy and
  Madeka}]{wen2017multihorizon}
\bibinfo{author}{Wen, R.}, \bibinfo{author}{Torkkola, K.},
  \bibinfo{author}{Narayanaswamy, B.}, \bibinfo{author}{Madeka, D.},
  \bibinfo{year}{2017}a.
\newblock \bibinfo{title}{A multi-horizon quantile recurrent forecaster}.
\newblock \URLprefix \url{https://arxiv.org/abs/1711.11053},
  \href{http://arxiv.org/abs/1711.11053}{{\tt arXiv:1711.11053}}.
\bibitem[{Wen et~al.(2017b)Wen, Torkkola, Narayanaswamy and
  Madeka}]{Wen2017-xz}
\bibinfo{author}{Wen, R.}, \bibinfo{author}{Torkkola, K.},
  \bibinfo{author}{Narayanaswamy, B.}, \bibinfo{author}{Madeka, D.},
  \bibinfo{year}{2017}b.
\newblock \bibinfo{title}{A {Multi-Horizon} quantile recurrent forecaster}, in:
  \bibinfo{booktitle}{Neural Information Processing Systems}.
\bibitem[{Ye and Dai(2021)}]{YE2021107617}
\bibinfo{author}{Ye, R.}, \bibinfo{author}{Dai, Q.}, \bibinfo{year}{2021}.
\newblock \bibinfo{title}{Implementing transfer learning across different
  datasets for time series forecasting}.
\newblock \bibinfo{journal}{Pattern Recognition} \bibinfo{volume}{109},
  \bibinfo{pages}{107617}.
\bibitem[{Zhang et~al.(2001)Zhang, Patuwo and Hu}]{zhang2001simulation}
\bibinfo{author}{Zhang, G.}, \bibinfo{author}{Patuwo, B.}, \bibinfo{author}{Hu,
  M.Y.}, \bibinfo{year}{2001}.
\newblock \bibinfo{title}{A simulation study of artificial neural networks for
  nonlinear time-series forecasting}.
\newblock \bibinfo{journal}{Computers \& Operations Research}
  \bibinfo{volume}{28}, \bibinfo{pages}{381 -- 396}.
\bibitem[{Zhang(2007)}]{zhang2007simulation}
\bibinfo{author}{Zhang, G.P.}, \bibinfo{year}{2007}.
\newblock \bibinfo{title}{A neural network ensemble method with jittered
  training data for time series forecasting}.
\newblock \bibinfo{journal}{Inf. Sci.} \bibinfo{volume}{177},
  \bibinfo{pages}{5329–5346}.
\newblock \DOIprefix\doi{10.1016/j.ins.2007.06.015}.
\bibitem[{Zhang and Qi(2005)}]{Zhang2005-le}
\bibinfo{author}{Zhang, G.P.}, \bibinfo{author}{Qi, M.}, \bibinfo{year}{2005}.
\newblock \bibinfo{title}{Neural network forecasting for seasonal and trend
  time series}.
\newblock \bibinfo{journal}{Eur. J. Oper. Res.} \bibinfo{volume}{160},
  \bibinfo{pages}{501--514}.
\bibitem[{Zhao and Itti(2018)}]{ZHAO2018171}
\bibinfo{author}{Zhao, J.}, \bibinfo{author}{Itti, L.}, \bibinfo{year}{2018}.
\newblock \bibinfo{title}{shapedtw: Shape dynamic time warping}.
\newblock \bibinfo{journal}{Pattern Recognition} \bibinfo{volume}{74},
  \bibinfo{pages}{171 -- 184}.
\bibitem[{Zhu and Laptev(2017)}]{Zhu2017IEEE}
\bibinfo{author}{Zhu, L.}, \bibinfo{author}{Laptev, N.}, \bibinfo{year}{2017}.
\newblock \bibinfo{title}{Deep and confident prediction for time series at
  uber}.
\newblock \bibinfo{journal}{2017 IEEE International Conference on Data Mining
  Workshops (ICDMW)} .

\end{thebibliography}
\endgroup

\end{document}